\documentclass[11pt,a4paper]{article}
\usepackage[hyperref]{emnlp2019}
\usepackage{times}
\usepackage{latexsym}

\usepackage{tikz, ifthen, amsmath}
\usepackage{subcaption}
\usepackage{caption}
\usepackage{array}
\usepackage{url}
\usepackage{booktabs}
\usetikzlibrary{calc, arrows, positioning, backgrounds, shapes.misc, math, matrix, shapes.callouts}

\definecolor{green_}{RGB}{179,226,205}
\definecolor{orange_}{RGB}{253,205,172}
\definecolor{blue_}{RGB}{203,213,232}
\definecolor{pink_}{RGB}{244,202,228}
\definecolor{lime_}{RGB}{230,245,201}
\definecolor{yellow_}{RGB}{255,242,174}

\newcolumntype{L}[1]{>{\raggedright\let\newline\\\arraybackslash\hspace{0pt}}m{#1}}
\newcolumntype{C}[1]{>{\centering\let\newline\\\arraybackslash\hspace{0pt}}m{#1}}
\newcolumntype{R}[1]{>{\raggedleft\let\newline\\\arraybackslash\hspace{0pt}}m{#1}}

\usepackage{xr-hyper}

\aclfinalcopy 

\title{Mastering emergent language: learning to guide in simulated navigation}

\author{Mathijs Mul \\
 University of Amsterdam \\
 {\tt mathijsmul@gmail.com} \\\And
  Diane Bouchacourt \\
  Facebook A.I. Research \\
  {\tt dianeb@fb.com} \\\And
  Elia Bruni \\
  Universitat Pompeu Fabra \\
  {\tt elia.bruni@gmail.com} \\}

\date{}

\begin{document}

\maketitle

\begin{abstract}
%
To cooperate with humans effectively, virtual agents need to be able to understand and execute language instructions. A typical setup to achieve this is with a scripted teacher which guides a virtual agent using language instructions. However, such setup has clear limitations in scalability and, more importantly, it is not interactive. 
Here, we introduce an autonomous agent that uses discrete communication to interactively guide other agents to navigate and act on a simulated environment. 
The developed communication protocol is trainable, emergent and requires no additional supervision. 
The emergent language speeds up learning of new agents, it generalizes across incrementally more difficult tasks and, contrary to most other emergent languages, it is highly interpretable.
We demonstrate how the emitted messages correlate with particular actions and observations, and how new agents become less dependent on this guidance as training progresses.
By exploiting the correlations identified in our analysis, we manage to successfully address the agents in their own language. 

\end{abstract}

\section{Introduction}
\label{sec:introduction}


Developing intelligent agents that can communicate with humans in a cooperative way is a longstanding goal of AI \citep{wooldridge2009mas, mikolov2016roadmap}.
Because supervised approaches are too static, not allowing agents to learn the interactive aspects of communication, current research on agent communication tends to focus on processes where language can naturally emerge \citep{lazaridou2017multiagent, mordatch2017emergence, havrylov2017emergence}.

In this paper, we introduce an autonomous agent (the Guide) that uses discrete communication to interactively assist another agent (the Learner) in navigational and operational tasks. In contrast with previous work where a Guide's language is hardcoded \citep{co2018guiding}, the communication protocol used by our Guide is fully emergent and does not require additional supervision to be trained. Nevertheless, the Guide speaks a language that preserves the desired properties of a manually programmed module: it is a language that generalizes compositionally across incrementally more difficult problems, and that is highly interpretable.



By analyzing the emergent communication protocols, we demonstrate strong correlations between messages and actions, and between messages and salient properties of the environment.
Quantifying the causal influence of communication shows how Learners that are assisted by a Guide gradually become less dependent on the message channel.
Moreover, we show how to interpret the ungrounded agent language, and even learn to `speak' it by letting an agent follow pre-set trajectories described by discrete messages.


The paper makes three main contributions. First, we introduce an autonomous agent that can interactively guide another agent via emergent, discrete communication, speeding up the learning in a collection of sequential decision making problems.
Second, we demonstrate that the developed language is general enough to be reused in levels different from the level it was developed on.
Third, we perform extensive analysis on the observed communication, which enables us to interpret the agents' language, and even to address them directly.

\section{Related Work}
\label{sec:related_work}


In this section we refer to important related work on the main themes of the current research: following language instructions, imitation learning and emergent communication.

\paragraph{Following language instructions}



Much research has been dedicated to linguistic instruction following. \citet{winograd1972understanding} tried to develop a program to interpret language directly, by hard-coding a great number of linguistic and physical regularities. More recently, \citet{cangelosi2006language} showed that robots equipped with neural networks are capable of learning action concepts from language instructions. Similar results were obtained by \citet{tellex2011understanding}, \citet{chen2011learning} and \citet{artzi2013weakly}.

In recent years, several artificial environments have been proposed to develop and assess agents' instruction-following capacities \citep{hermann2017grounded, brodeur2017home, chaplot2018gated, wu2018building}. 
Relevant research includes \citet{mei2016listen}, \citet{yu2018interactive}, \citet{bahdanau2018learning}, \citet{wang2016learning} and \citet{williams2018learning}.


In this work we use a synthetic gridworld with natural language instructions called `BabyAI'. \citet{chevalier2018babyai} introduced the set-up and presented a baseline model. 
\citet{co2018guiding} extended this model with a correction module, which provides additional linguistic feedback to an agent. This module, however, is fully hard-coded. We introduce a comparable mechanism that is entirely trained.

\paragraph{Imitation learning}


Most of the models in our research are trained using imitation learning, where the objective of an agent is to simulate the behavior of another agent with more experience or knowledge of a particular task.
Forms of imitation learning have been successfully applied in the context of following language instructions by e.g. \citet{hemachandra2015learning}, \citet{bahdanau2018learning} and \citet{chevalier2018babyai}.


A range of imitation learning algorithms has been proposed in the literature, including Searn \citep{daume2009search}, SMILe \citep{ross2010} and DAgger \citep{ross2011reduction}.
Inverse reinforcement learning can also be applied, as suggested by \citet{irl1}, \citet{irl2} and \citet{ziebart2008maximum}.
We use behavioral cloning \citep{behavcloning}, which treats expert trajectories as series of states labelled with actions that a new agent can learn to predict in a supervised setting.



\paragraph{Emergent communication}



In the current research, we study the emergence of communication through discrete language in an artificial, multi-agent set-up. The emergence and evolution of language has received much attention in linguistics, game theory and cognitive science \citep{skyrms2010, wagner2003progress, crawford1998survey}. 

Experiments related to ours were conducted by \citet{jorge2016learning} and \citet{sukhbaatar2016learning}, who showed that agents can autonomously evolve a communication protocol that helps them to play a game. \citet{lazaridou2017multiagent} obtained similar results, but also investigated how to change the environment in order to interpret the emergent language more easily. Inspired by this research, \citet{havrylov2017emergence} developed a referential game where agents learn to assist each other by sending variable-length sequences of discrete symbols. \citet{mordatch2017emergence} showed that multi-agent communities can give rise to a grounded compositional language that helps speakers achieve their goals.  


%
%

\section{Approach}
\label{sec:approach}

In this section we describe the BabyAI game, as well as the models used in the experiments and the training regime.

\subsection{BabyAI game}
\label{sec:babyai}

\begin{figure}[h]
\centering
\includegraphics[width=0.25\textwidth]{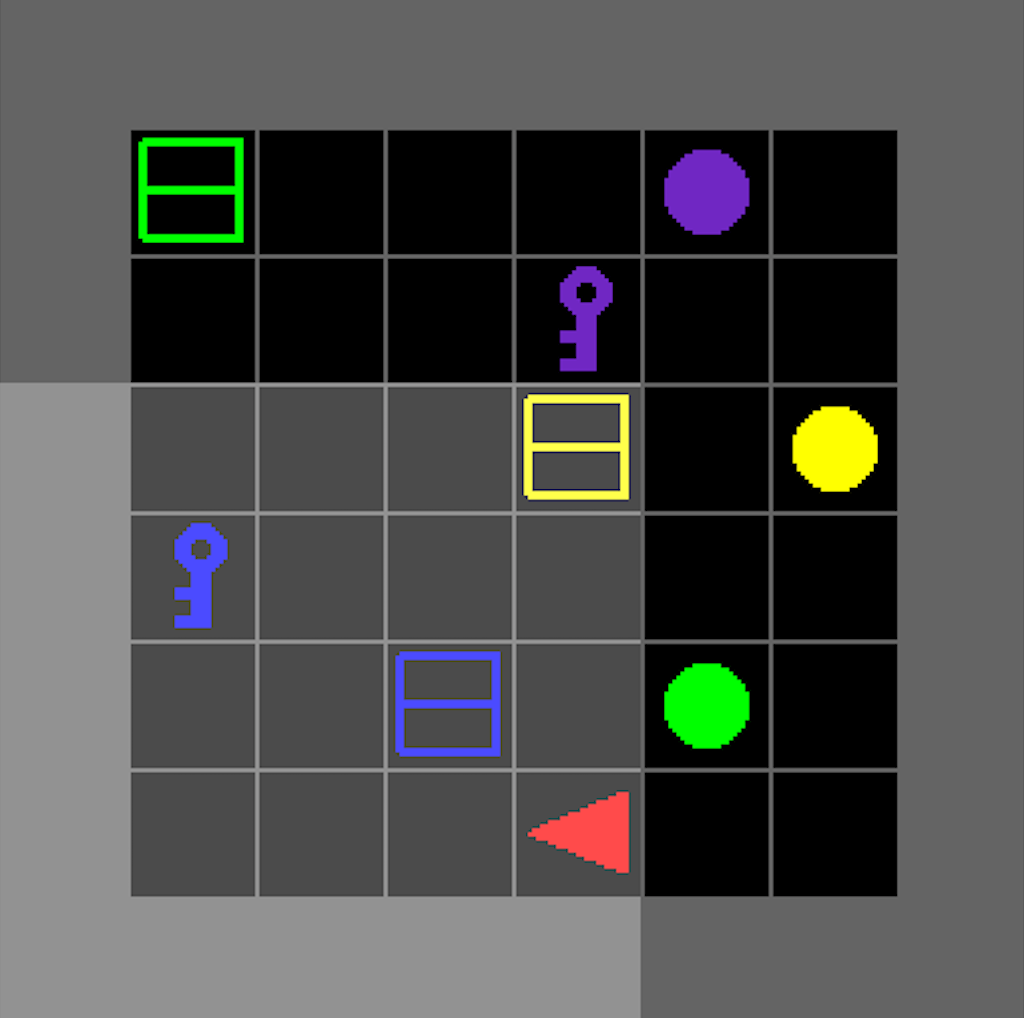}
\caption{A screenshot of BabyAI level PutNextLocal with mission `put the green box next to the green ball'. The red arrow represents the agent and the shaded area the currently observable part of the environment.}
\label{fig:babyai-screenshot}
\end{figure}

All our experiments are conducted using the BabyAI framework, introduced by \citet{chevalier2018babyai}.\footnote{Code available at https://github.com/mila-udem/babyai.} The platform lets virtual agents play games at a number of different levels. Each of these levels are combinations of a partially observable two-dimensional gridworld, and a mission presented as a language instruction. The challenges include various operational and navigational tasks. The environment contains several kinds of objects (e.g. balls, keys, boxes and doors) that come in different colors. See Figure \ref{fig:babyai-screenshot} for a sample frame.

At each time step $t$, an agent receives as input the instruction string $i_t$ (constant during a game), and the $7 \times 7 \times 3$-tensor $o_t$ encoding the currently observable part of the surroundings. Her subsequent action $a_t$ alters the state of the environment and thereby determines the next observation $o_{t+1}$.

In total, BabyAI has 19 levels with increasing degrees of complexity. In this paper, we consider six of these levels, which we consider to be a representative fragment of the game as a whole. See Table A.1 in the supplementary materials for an overview of the specific core skills required to solve each of them. The BabyAI levels are designed in such a way that competencies required to solve some of them are also needed in others. E.g., to complete GoToLocal an agent has to know how to navigate a single room with distractors, and this skill is also necessary to address PickupLoc. Thanks to this hierarchy, BabyAI is a suitable framework for studying curriculum learning.

%
%
%
%
%
%
%

\subsection{Model}

We consider two kinds of agents: \textit{Learners} and \textit{Guides}. The fully continuous Learners can approach a task alone, or be assisted by a Guide. A Guide is an agent that not only learns to solve a task, but also to emit discrete messages. These messages can be used to help a new Learner master a task more efficiently.\footnote{Code will be released upon acceptance.}

\paragraph{The Learner}

The Learner architecture is as described by \citet{chevalier2018babyai}. Instructions are presented to a GRU \citep{cho2014gru} and observations to a convolutional network. The results of these operations are processed together by a FiLM module \citep{perez2018film} to obtain a joint representation of the command and the environment. A memory LSTM cell \citep{hochreiter1997lstm} uses this representation, together with past FiLM outputs, to compute the tensor that is passed to a two-layered policy module.
The action receiving the highest probability after applying softmax to the policy output is executed.

%
%
%



\paragraph{The Guide}

The Guide's architecture is similar to the Learner's, but differs in one crucial aspect: between the memory LSTM and the policy module we introduce a bottleneck.
Its function is to create the communication channel, where the information flowing from the input-processing layers has to be compressed in the form of a sequence of discrete tokens. These messages can later be sent to a Learner.

Generation of the messages takes place in a GRU decoder that receives the Guide's memory LSTM output as input. 
To do so, continuous information has to be discretized.
In order to maintain end-to-end differentiability, we use a straight-through Gumbel softmax estimator for this purpose  \citep{bengio2013estimating, jang2017categorical}.
The resulting string of symbols can be passed to the policy module to predict an action after it has been encoded again by a GRU encoder.

%
%
%
%

%



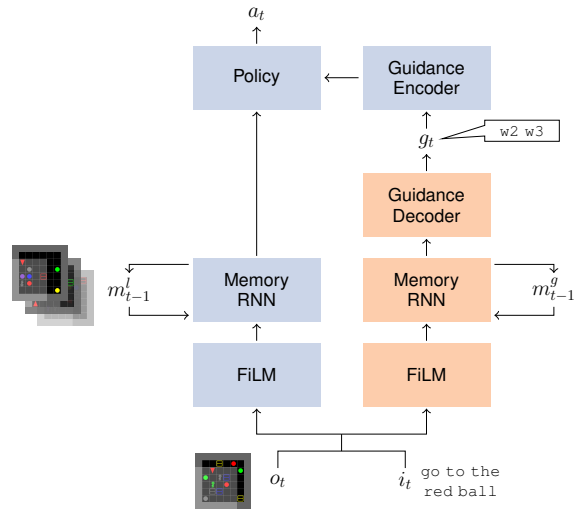
\begin{figure}[h]
	\centering
	\begin{tikzpicture}[scale=0.6, 
		every node/.style={scale=0.6},
		module/.style={rectangle, minimum width=3cm, minimum height=1.5cm},
		var/.style={rectangle, minimum width=1cm, minimum height=1cm},
		notice/.style  = { draw, rectangle callout, callout relative pointer={#1}}]
		
		\def\nodedist{2cm};
		\pgfdeclarelayer{bg}    
		\pgfsetlayers{bg,main}  

		\node[module, fill=blue_] (pi) at (0, 5) {\large \textsf{Policy}};
		\node[module, fill=blue_, align=center] (memrnn) at (0,0) {\large \textsf{Memory} \\ \large \textsf{RNN}};
		\node[module, fill=blue_] (film) at (0,-2) {\large \textsf{FiLM}};
		\node[module, fill=blue_, align=center] (correncoder) at (4, 5) {\large \textsf{Guidance} \\ \large \textsf{Encoder}};
		\node[rectangle, fill=white, align=center] (correction) at (-3, 0) {\Large $m^l_{t-1}$};

		\node[rectangle, fill=white, align=center] (correction) at (0, 6.5) {\Large $a_t$};

		\path [->,draw] (0, -1.2) -- (0, -0.8) ; 
		\path [->,draw] (0, 0.8) -- (0, 4.2) ; 

		\path [->, draw] (-1.6, 0.6) -- (-3, 0.6) -- (-3, 0.4);
		\path [<-, draw] (-1.6, -0.6) -- (-3, -0.6) -- (-3, -0.4);

		\path [->, draw] (0, 5.8) -- (0, 6.2); 


		\node[module, fill=orange_, text centered, align=center] (memrnn) at (4,0) {\large \textsf{Memory}\\ \large \textsf{RNN}}; 
		\node[module, fill=orange_] (film) at (4,-2) {\large \textsf{FiLM}};
		\node[module, fill=orange_, align=center] (corrdecoder) at (4, 2) {\large \textsf{Guidance} \\ \large \textsf{Decoder}};

		\node[rectangle, fill=white, align=center] (correction) at (4, 3.5) {\Large $g_t$};

		\path [->,draw] (4, -1.2) -- (4, -0.8) ; 
		\path [->,draw] (4, 0.8) -- (4, 1.2) ; 
		\path [->, draw] (4, 2.8) -- (4, 3.2); 
		\path[->, draw] (2.4, 5) -- (1.6, 5); 
		\path[->, draw] (4, 3.8) -- (4, 4.2); 

		\node[rectangle, fill=white, align=center] (memcorr) at (7, 0) {\Large $m^g_{t-1}$};

		\path [->, draw] (5.6, 0.6) -- (7, 0.6) -- (7, 0.4);
		\path [<-, draw] (5.6, -0.6) -- (7, -0.6) -- (7, -0.4);

		\node[rectangle, fill=white, align=center] (correction) at (0.5, -4.5) {\Large $o_t$};
		\node[rectangle, fill=white, align=center] (correction) at (3.5, -4.5) {\Large $i_t$};

		\path [draw] (0.5, -4.2) -- (0.5, -3.8) -- (3.5, -3.8) -- (3.5, -4.2);
		\path[draw] (2, -3.8) -- (2, -3.4);
		\path[draw, ->] (2, -3.4) -- (0, -3.4) -- (0, -2.8);
		\path[draw, ->] (2, -3.4) -- (4, -3.4) -- (4, -2.8);

		\node (obs-now) at (-0.8, -4.5) {\includegraphics[scale=0.2]{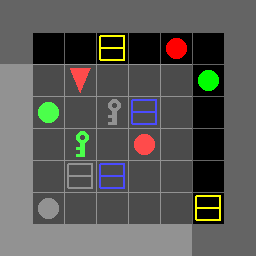}};
		\node[text width=2cm, align=center] (ins-now) at (4.8, -4.5) {\texttt{go to the red ball}};
		\begin{pgfonlayer}{bg}
    			\node[opacity=0.4] (mem-3) at (-4.5,-0.2) {\includegraphics[scale=0.2]{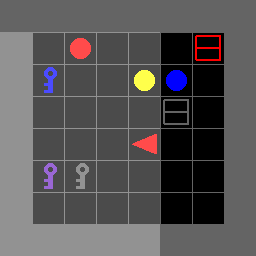}};
    			\node[opacity=0.8] (mem-2) at (-4.8,0.1) {\includegraphics[scale=0.2]{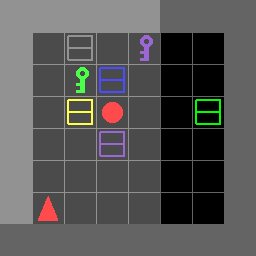}};
    			\node (mem-1) at (-5.1,0.4) {\includegraphics[scale=0.2]{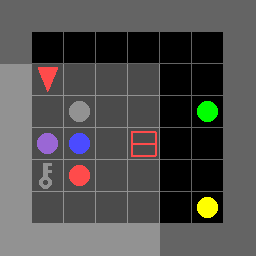}};
		\end{pgfonlayer}

		\node[notice={(-1,-0.1)}, text width=1.5cm, align=center] at (6.25,3.75) {\texttt{w2 w3}};
	\end{tikzpicture}
\caption{Schematic visualization of the model architecture. Included variables are $o_t$: state of the (observable part of the) environment, $i_t$: linguistic instruction, $g_t$: linguistic guidance, $a_t$: action, $m^l_{t-1}$: Learner's memory state, $m^g_{t-1}$: Guide's memory state. Subscript represents point in time. Blue modules are considered part of the Learner; red ones part of the Guide.}
    \label{fig:model}
\end{figure}

By first training a Guide and then pairing the experienced Guide with a new Learner, we investigate if and how agents can learn to make use of the available communication channel. The set-up used to test this is visualized in Figure \ref{fig:model}. A new Learner is coupled with a pretrained Guide, and receives her discrete guidance at each time step. The Learner is provided with a new GRU encoder to interpret the guidance, and uses this information as additional input to her policy module.

\subsection{Training}

Because we use behavioral cloning as imitation learning algorithm, we effectively reduce training to a supervised setting.
The Learner, the Guide and the combined Learner-Guide set-ups are all end-to-end architectures that receive observations and instructions as input, and whose parameters are updated by backpropagating a negative log likelihood loss function.

%
%
%
%

The agents are supposed to internalize the policy demonstrated by an expert agent, which requires that we first have such an expert at our disposal. \citet{chevalier2018babyai} implemented a heuristic bot that is able to solve all BabyAI levels. We do not use this hard-coded expert, but instead apply proximal policy optimization \citep{schulman2017ppo} to train RL (Reinforcement Learning) experts. We do this because it makes our approach more generalizable, and because \citeauthor{chevalier2018babyai} note that `demonstrations produced by [an RL] agent are easier for the learner to imitate' than those produced by a symbolic bot. Therefore, our experiments follow three consecutive training stages: (1) Train RL expert (with Learner set-up) to generate demonstrations.
(2) Train new Learner (as baseline) and new Guide separately on the RL expert data, by imitation learning. (3) Train new Learner + pretrained Guide (from stage 2) on the RL expert data, by imitation learning.
In stage 3, the pretrained Guide's weights are not frozen so they can be finetuned. 

%
%
%
%
%

Stage 1 only has to be performed once to generate the expert data that we use in imitation learning.
In all our experiments, the Guide emits messages of a fixed length of 2. There are three distinct tokens (\texttt{w0}, \texttt{w1} and \texttt{w2}), so that there are 9 possible messages in total.
For an overview of the hyperparameter settings we refer to Table A.2 in the supplementary materials.

\section{Experiments and results}
\label{sec:experiments}

\begin{figure*}[t]
\centering
\begin{subfigure}{0.32\textwidth}
\includegraphics[width=\textwidth]{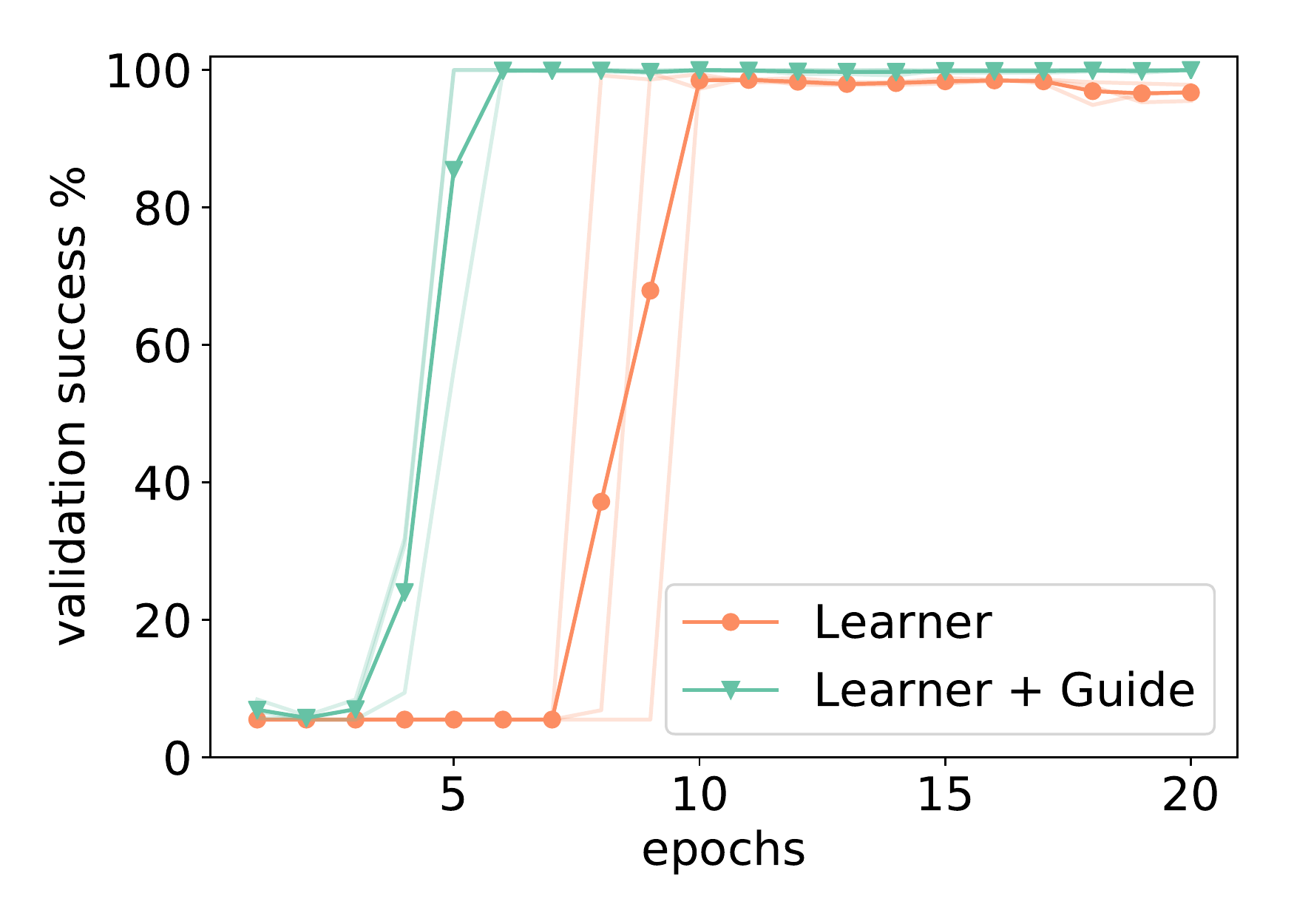}
\caption{GoToObj}
\end{subfigure}
\begin{subfigure}{0.32\textwidth}
\includegraphics[width=\textwidth]{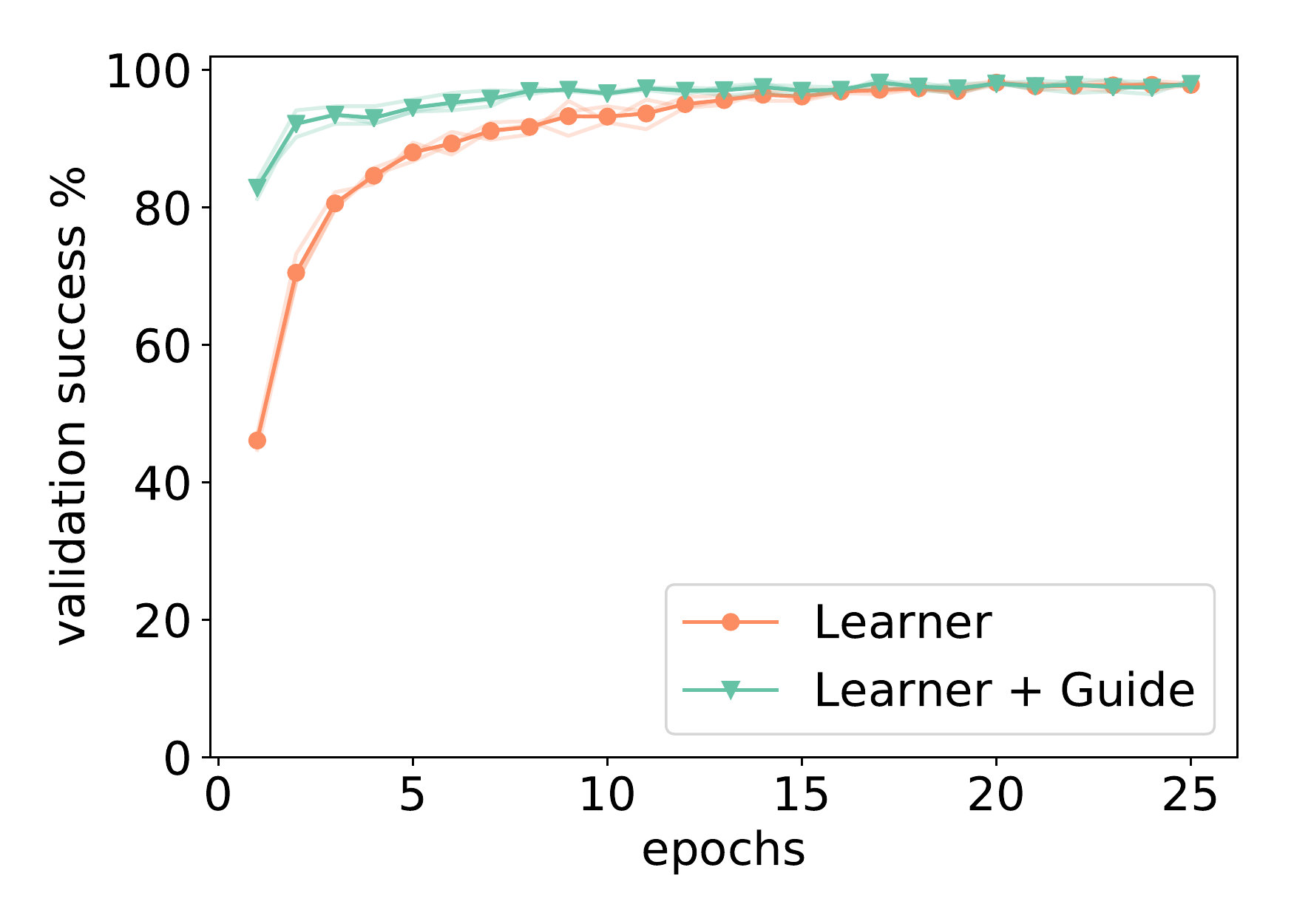}
\caption{GoToLocal}
\end{subfigure}
\begin{subfigure}{0.32\textwidth}
\includegraphics[width=\textwidth]{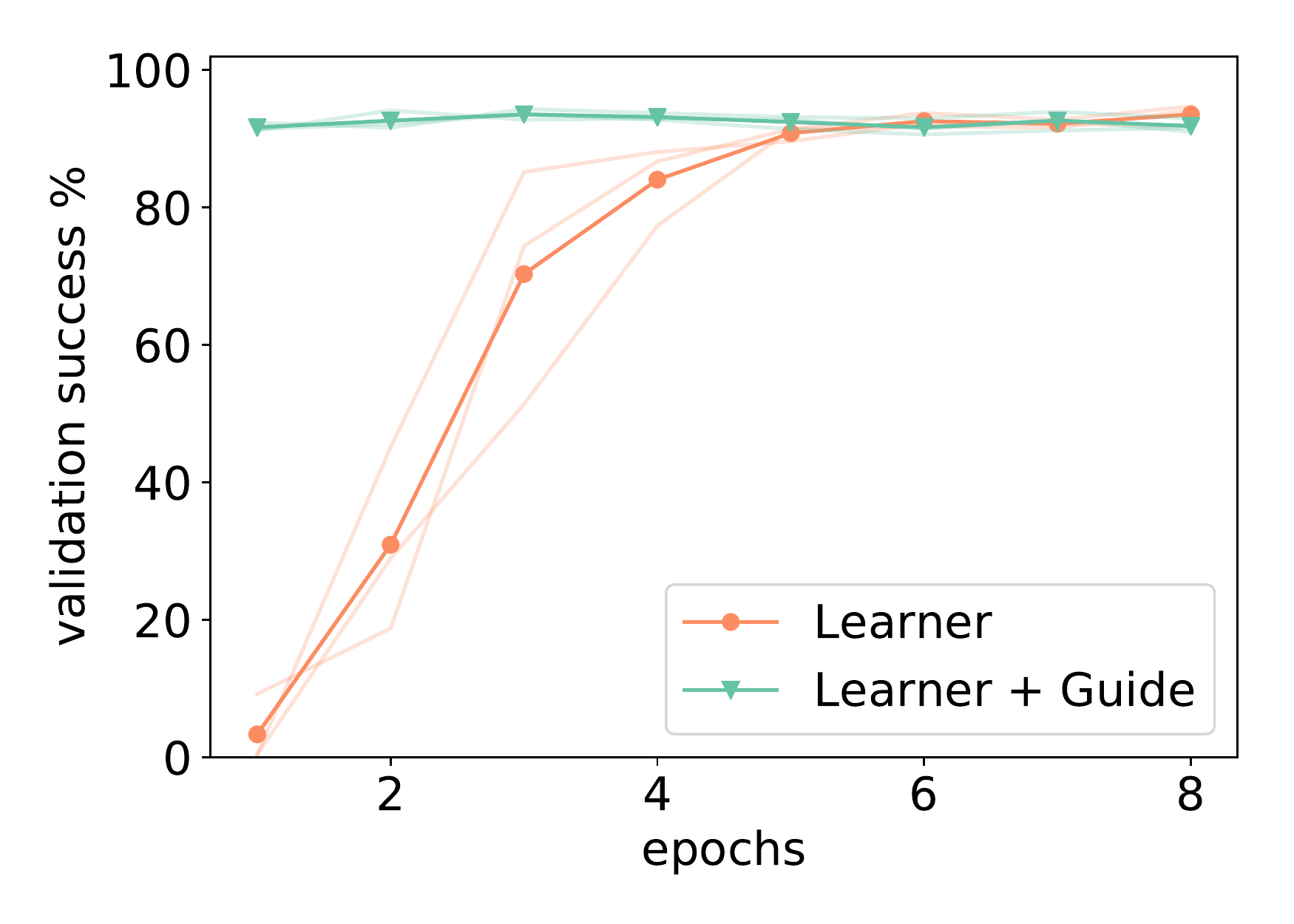}
\caption{GoToObjMaze}
\end{subfigure}
\begin{subfigure}{0.32\textwidth}
\includegraphics[width=\textwidth]{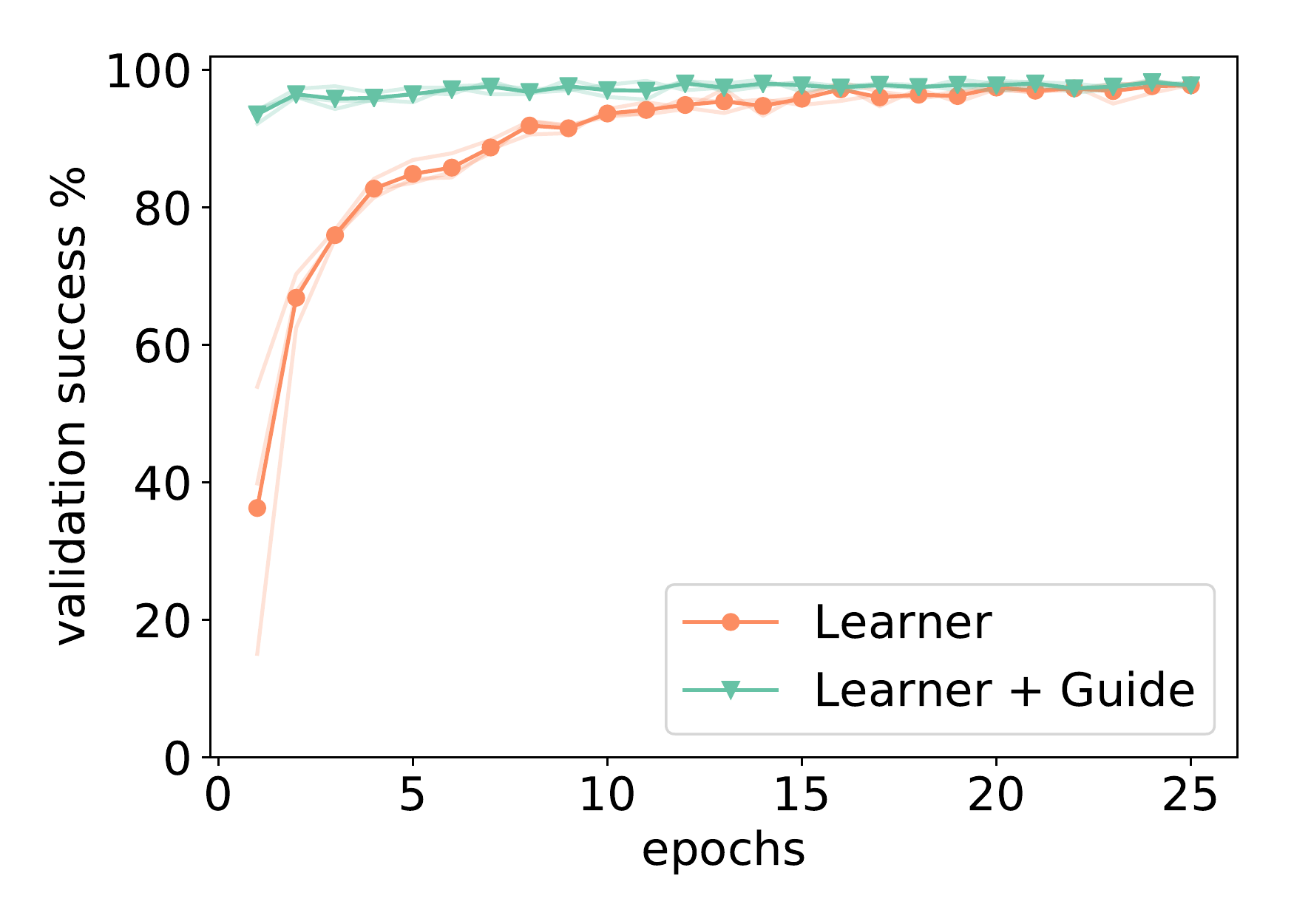}
\caption{PickupLoc}
\end{subfigure}
\begin{subfigure}{0.32\textwidth}
\includegraphics[width=\textwidth]{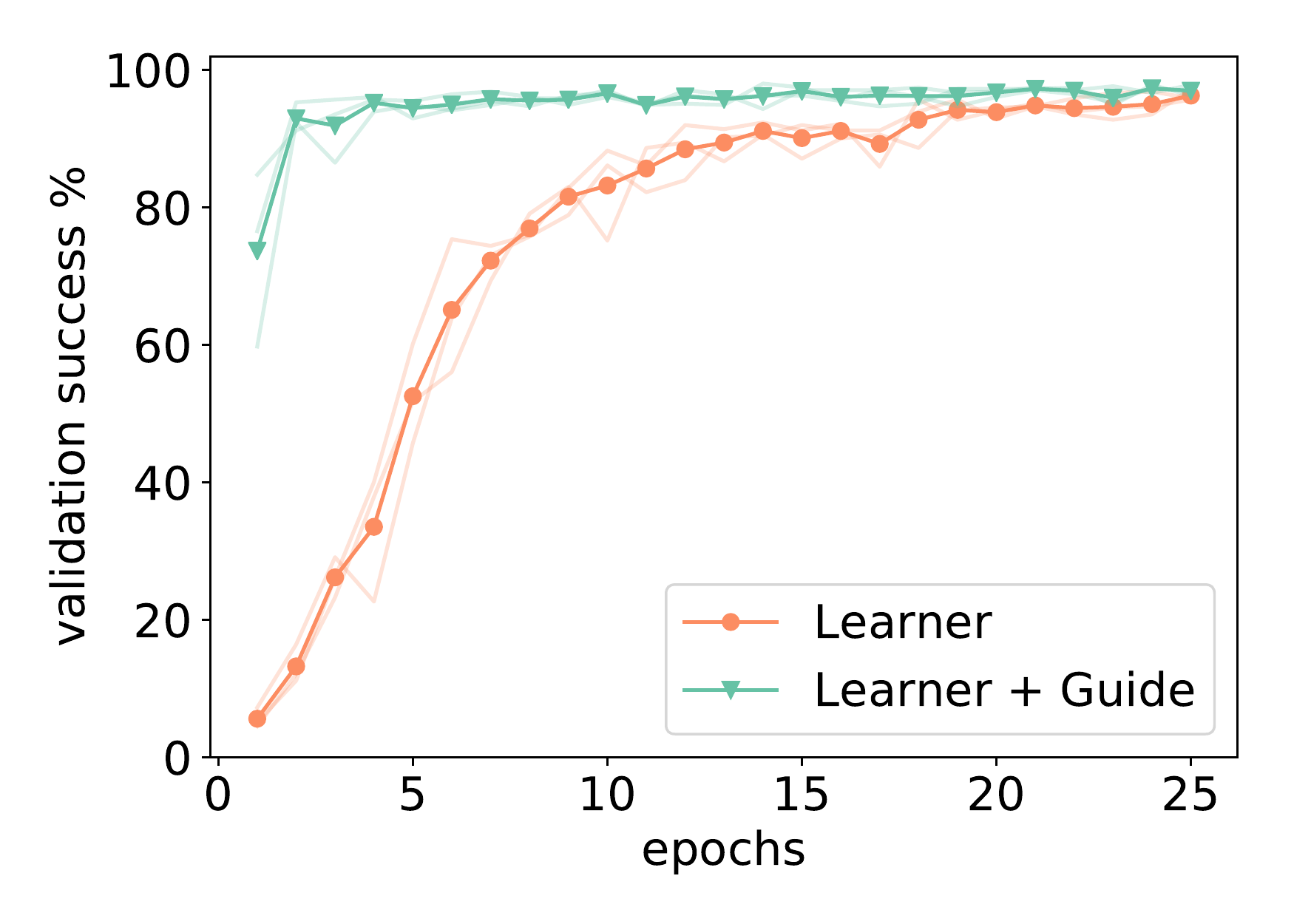}
\caption{PutNextLocal}
\end{subfigure}
\begin{subfigure}{0.32\textwidth}
\includegraphics[width=\textwidth]{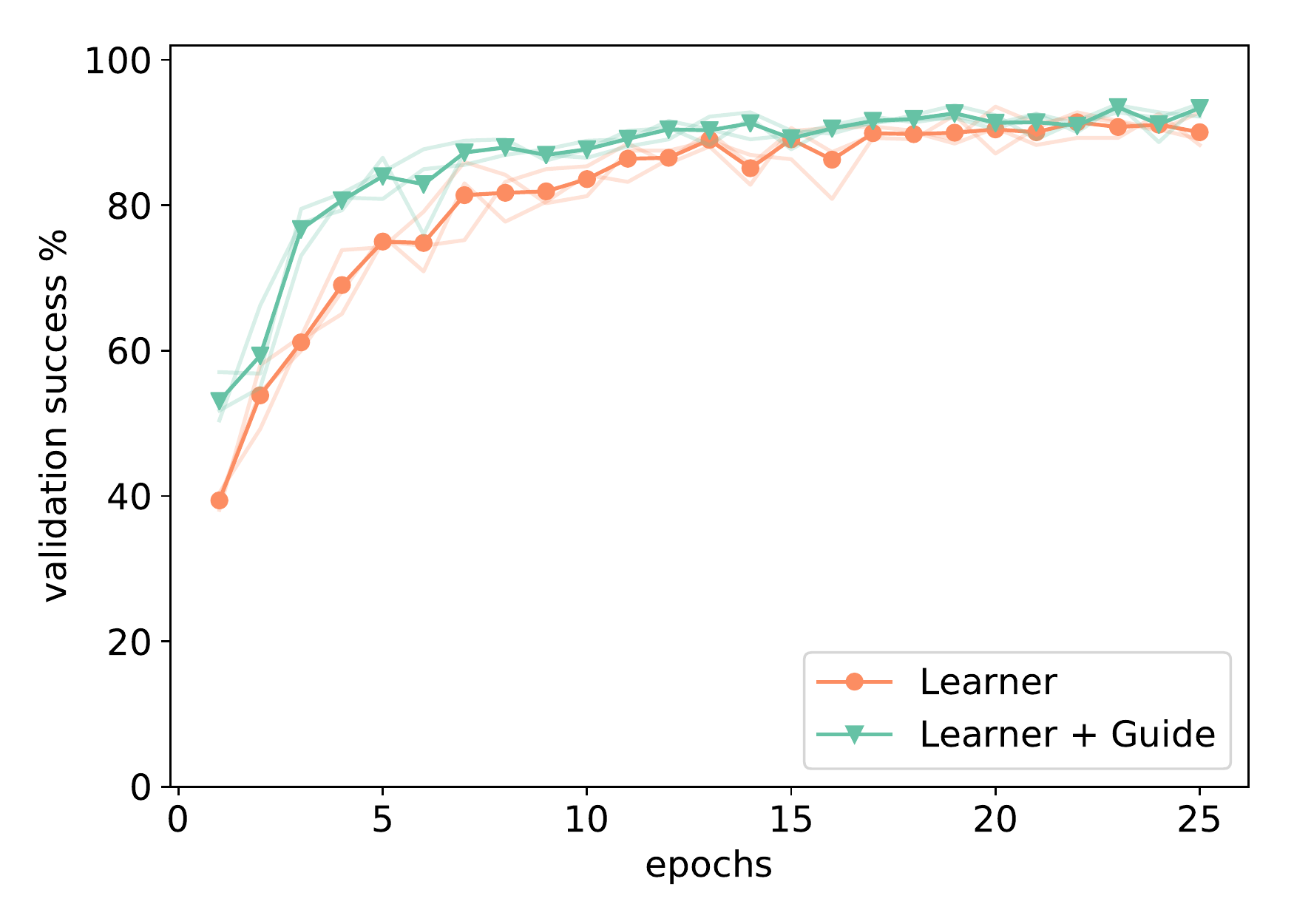}
\caption{GoTo}
\end{subfigure}
\caption{Development of validation success rate when training a single Learner and a Learner assisted by a pretrained Guide on the indicated BabyAI levels. Average over three runs; results per run are shown by shaded lines.}
\label{fig:learner-guide-intralevel}
\end{figure*}

\subsection{Intra-level guidance}

First, we study the effect of the communication channel within single levels. We train the baseline model (the default Learner), and compare its performance over time with a Learner that receives messages from the best-performing pretrained Guide, which is itself being finetuned to coadapt with the Learner. We compare the models in terms of validation success rate, which is the percentage of validation episodes (out of 500) that a model manages to complete successfully. Note that this metric differs from the accuracy, which is the percentage of individual actions that are predicted correctly.

We do not expect the guided Learner to achieve a higher success rate than the baseline Learner, because we already know that the Learner is in principle able to solve all levels when given enough training time. What we do hope to see is that a new Learner masters a task more quickly when aided by the messages of a pretrained Guide. 

Results of three runs with three random seeds on all levels are shown in Figure \ref{fig:learner-guide-intralevel}. These plots clearly show that on each of the considered levels, the messages emitted by the pretrained Guide cause a great increase in the learning speed of a new agent.



\subsection{Curriculum guidance}

%

We assess if the Guide is also useful when we apply curriculum learning: are the messages emitted by a Guide pretrained at one level helpful for a Learner facing a more complex level? We consider transfer from GoToLocal to PickupLoc and PutNextLocal, and between PickupLoc and PutNextLocal (both directions).

\begin{figure*}[t]
\centering
\begin{subfigure}{0.49\textwidth}
\includegraphics[width=\textwidth]{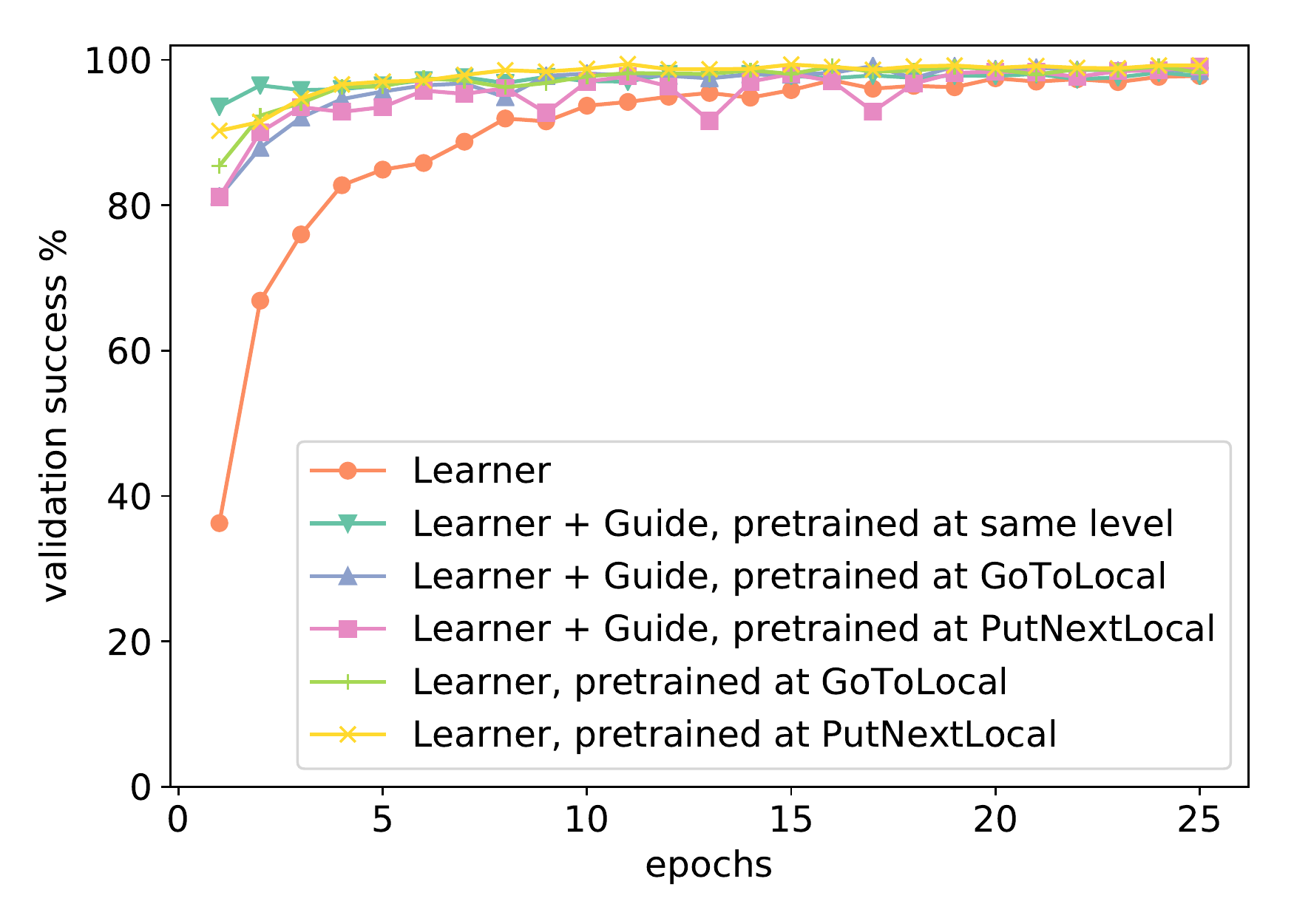}
\caption{PickupLoc}
\end{subfigure}
\begin{subfigure}{0.49\textwidth}
\includegraphics[width=\textwidth]{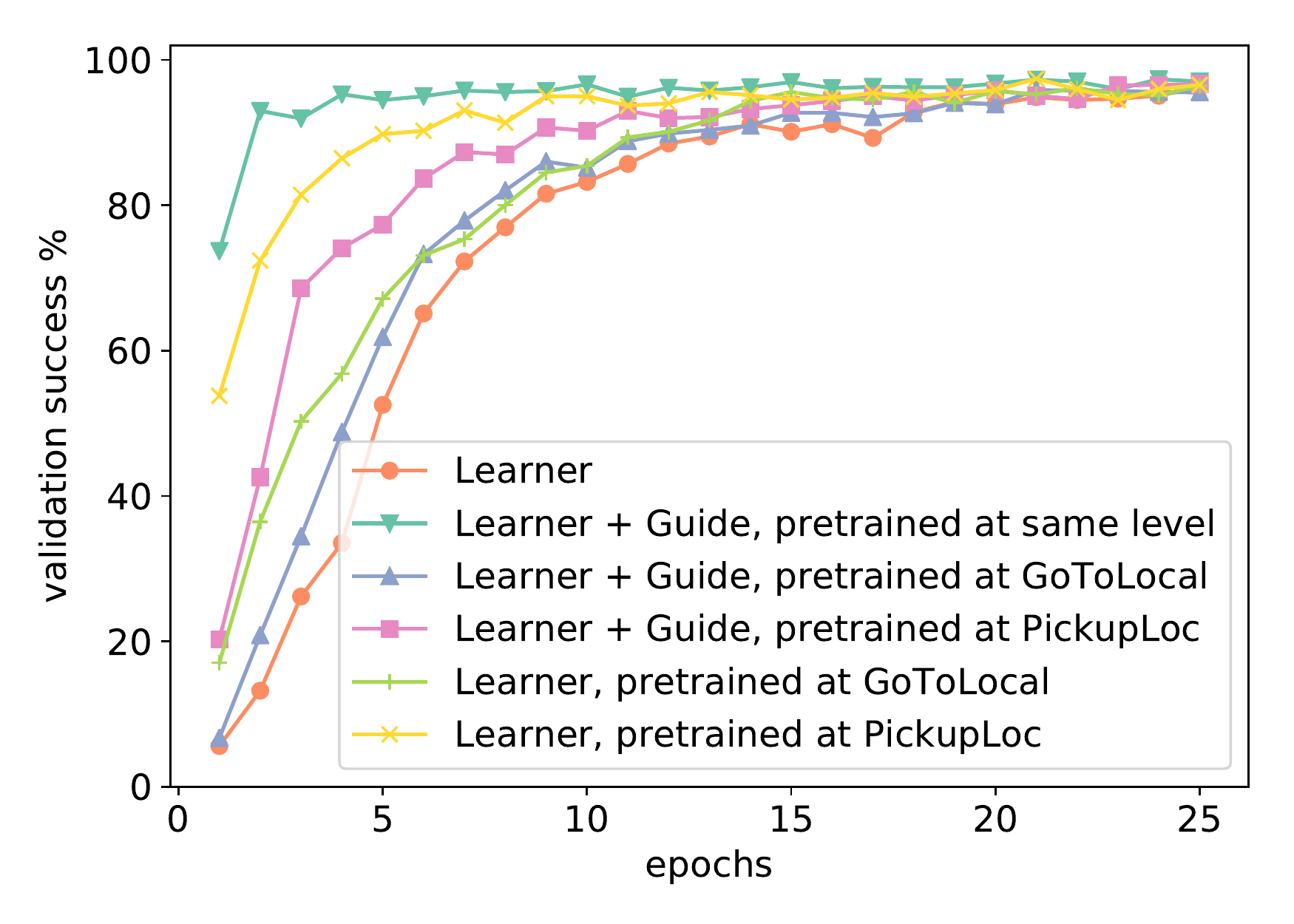}
\caption{PutNextLocal}
\end{subfigure}
\caption{Development of validation success rate when training a single Learner from scratch, a Learner assisted by a pretrained Guide, or a pretrained Learner on several BabyAI levels.
}
\label{fig:learner-guide-interlevel}
\end{figure*}

%
Guides are trained at the mentioned base levels, and coupled with a new Learner at the related target levels. In all of these cases, we hypothesize that the messages sent by the Guide should be useful also at the target levels, even though the Guide never actually encountered them. This is because all of the levels concerned require an agent to be capable of navigating a single room with distractors (competencies ROOM, DISTR-BOX and DISTR of Table A.1 in the supplementary material). Hence, if the guidance messages are sufficiently general, they should be useful across base and target levels.

We compare performance of the models with different base and target levels with those for which base and target level coincide (discussed in the previous section). Additionally, we assess the performance at the target levels of a single Learner pretrained at the base levels. We expect a pretrained Learner to perform better than a new Learner assisted by a pretrained Guide, because in the former all parameters have been optimized for a related level, whereas in the latter the Learner can only profit from the pretraining by means of the discrete guidance messages.

Figure \ref{fig:learner-guide-interlevel} shows the results. In level PickupLoc, we see that pretraining the Guide at levels GoToLocal and PutNextLocal brings the learning curve close to the one of a learner with a Guide pretrained at PickupLoc itself. Surprisingly, the difference between the new Learners with a pretrained Guide and the pretrained Learners is not significant. In PutNextLocal, the results are more divergent, and the Learner with a Guide pretrained at the same level outperforms all other set-ups. Especially pretraining the Guide at PickupLoc accelerates learning, but not as much as pretraining the Learner at PickupLoc.

The observation that Guides can even be helpful to new Learners at levels for which they were not optimized suggests that their messages convey information that is relevant to the internalization of skills that generalize across levels.


\subsection{Multi-level guidance}

\begin{figure*}[t]
\centering
\begin{subfigure}{0.32\textwidth}
\includegraphics[width=\textwidth]{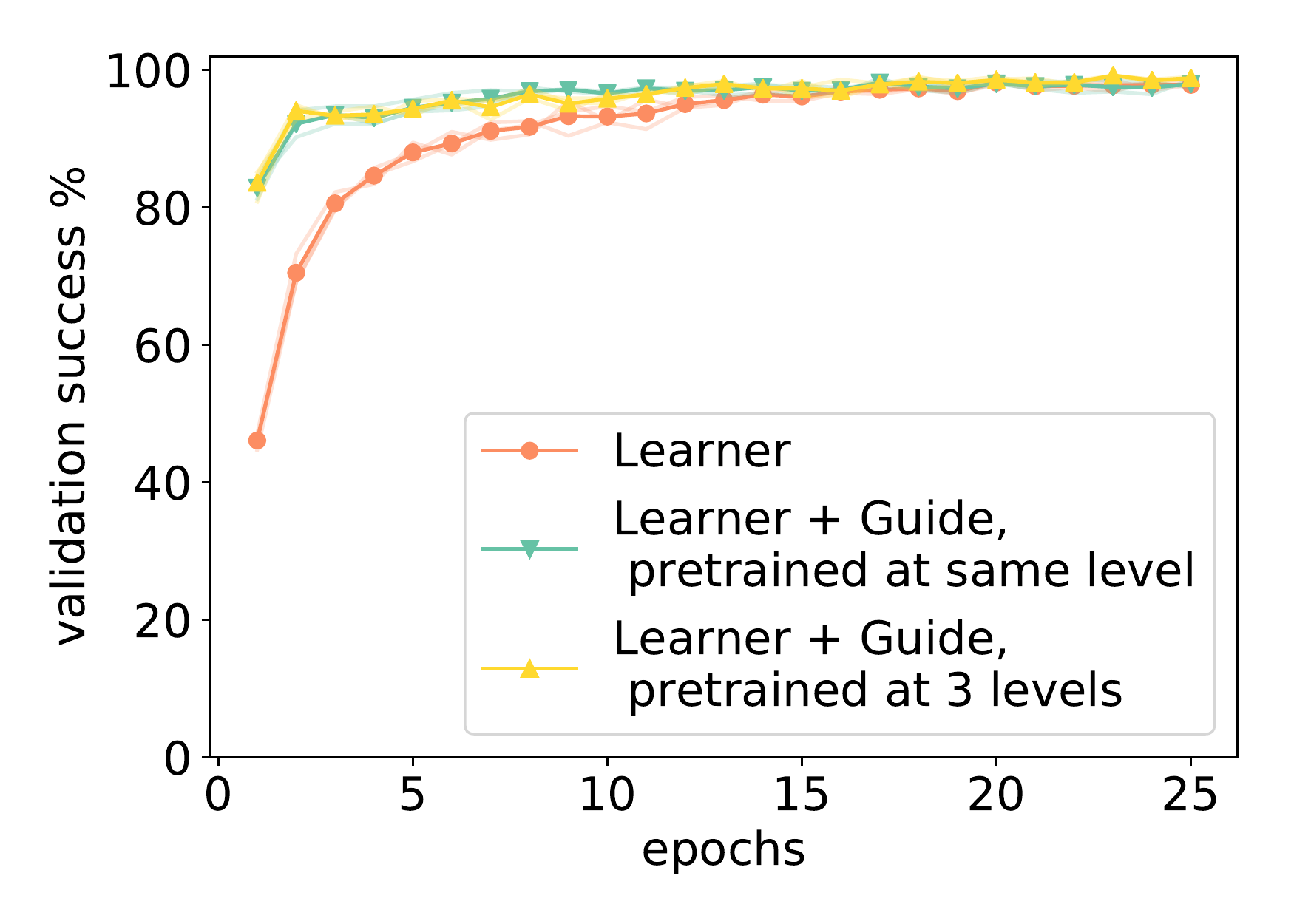}
\caption{GoToLocal}
\end{subfigure}
\begin{subfigure}{0.32\textwidth}
\includegraphics[width=\textwidth]{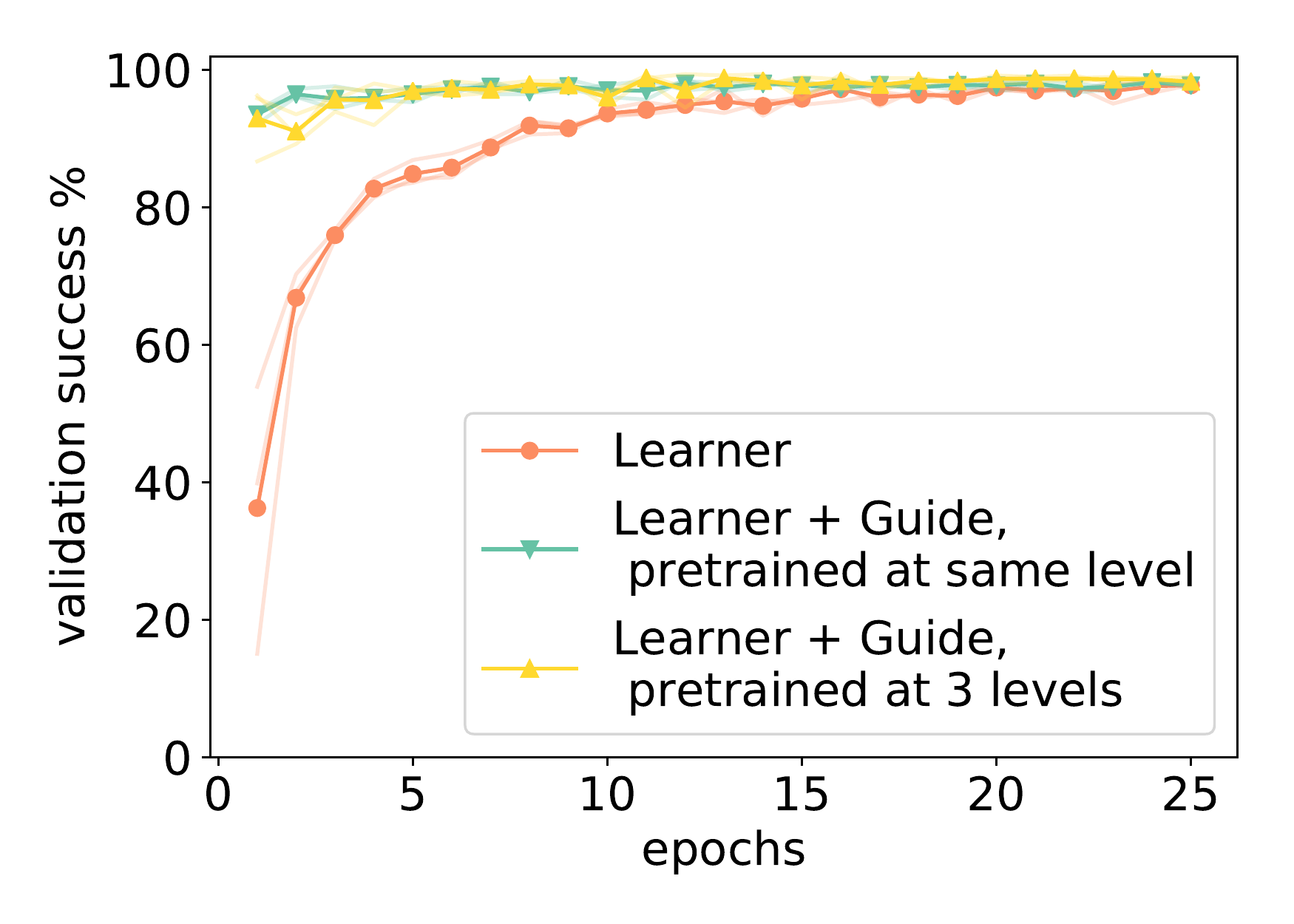}
\caption{PickupLoc}
\end{subfigure}
\begin{subfigure}{0.32\textwidth}
\includegraphics[width=\textwidth]{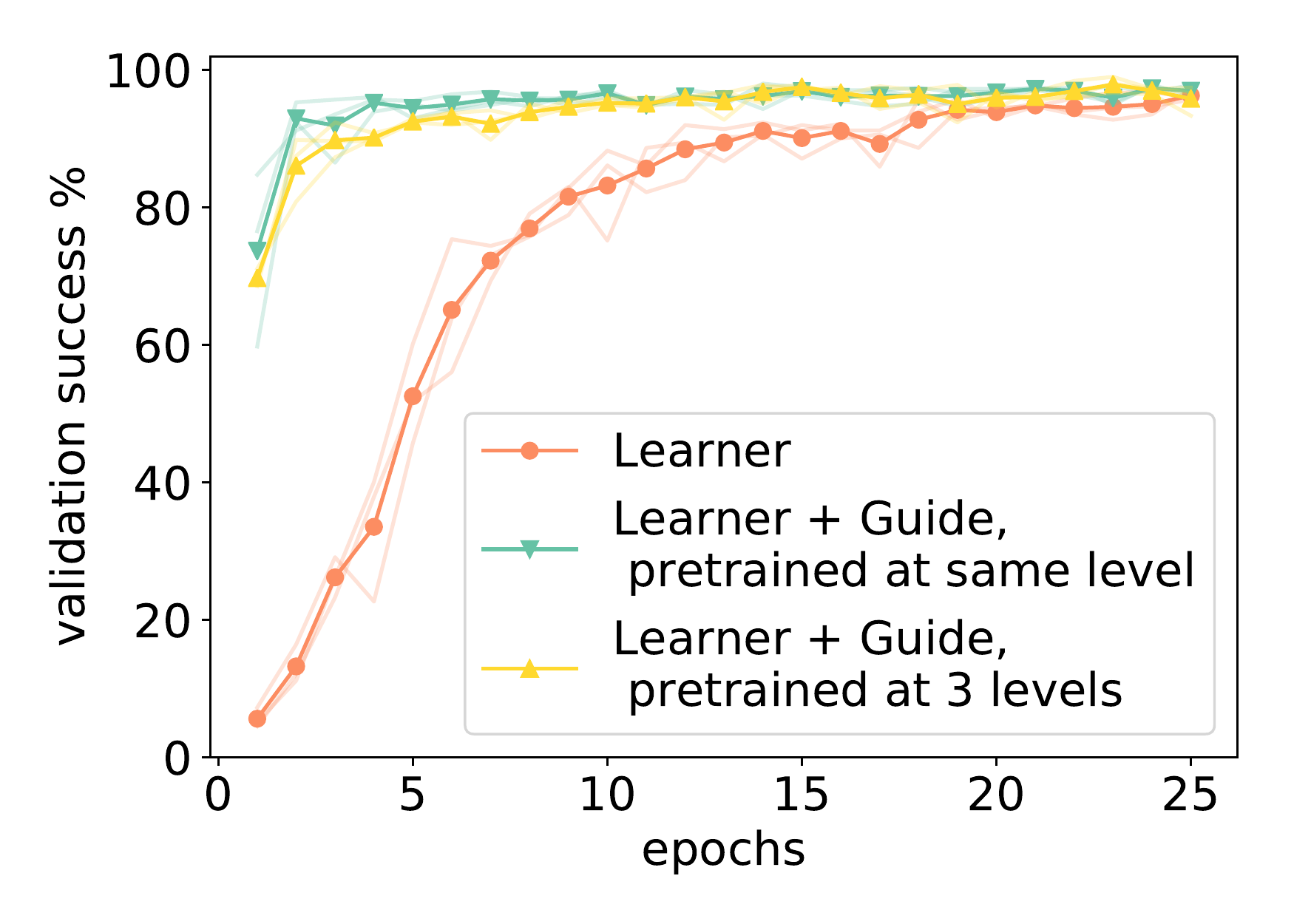}
\caption{PutNextLocal}
\end{subfigure}
\caption{Development of validation success rate when training a single Learner, a Learner assisted by a Guide pretrained on a single level and a Learner pretrained on GoToLocal, PickupLoc and PutNextLocal simultaneously. Average over three runs; results per run are shown by shaded lines.}
\label{fig:learner-multiguide}
\end{figure*}

The previous experiment focused on the transfer of guidance from one level to another one. This transfer can be considered inductive because it relies on generalization to unseen levels. In the next experiment we also check for a deductive transfer: if a Guide is trained on a combination of levels, how useful is she when assisting a Learner trained from scratch on only one of these levels?

To investigate this, we train a single Guide on GoToLocal, PickupLoc and PutNextLocal at the same time, by aggregating the training sets of these three levels. None of the model dimensions are changed, implying that this multi-level Guide has to address three different levels with the same number of parameters as a regular Guide. Because of this handicap, we hypothesize that the multi-level Guide will not accelerate the development of a new Learner as rapidly as a Guide specialized at a single level.

The results of this experiment are visualized in Figure \ref{fig:learner-multiguide}. At all three levels the multi-level Guide turns out to be just as helpful to a new Learner as a Guide that was specifically optimized for a single level. This again suggests that the Guide does not produce messages aimed at solving specific levels, but rather at acquiring skills that are relevant in each of them.


\section{Analysis}
\label{sec:analysis}

In this section we investigate the communication evolved by the agents in more detail. 

\subsection{Messages/actions correlations}
\label{sec:messages-actions}

\begin{figure}[h]
\includegraphics[width=0.45\textwidth, trim={0 0 0 1cm}, clip]{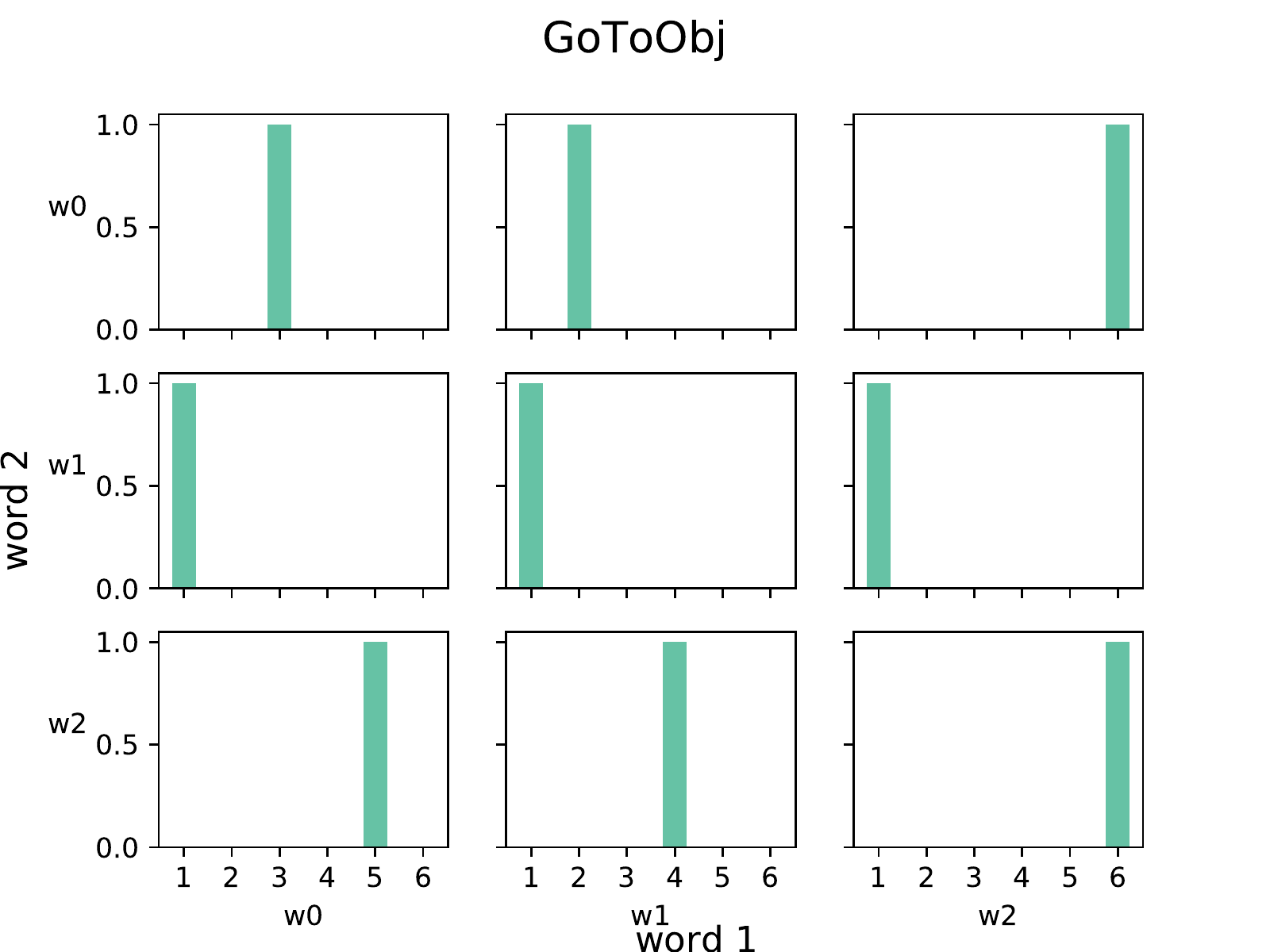}
\caption{Barplots visualizing the conditional distribution of actions taken in PutNextLocal by the best Guide after stage 2, given the generated messages, based on 500 sample episodes unseen during training. The bins in the individual barplots correspond to actions and the bar heights to probabilities. E.g. the top left barplot shows that the Guide performs action 3 with probability 1.0 after emitting message \texttt{w0 w0}.}
\label{fig:action-message-dist-stage2-putnextlocal}
\end{figure}

%

We analyse the conditional probability distribution of actions given the messages produced by the pretrained or finetuned Guide. This reveals a strong correlation between messages and actions, as illustrated in Figure A.1, visualizing the conditional distribution of actions taken by the best-performing Guide, conditioned on her own messages, after training stage 2.

We show the results for level PutNextLocal, because here the agent has to perform most actions. Similar plots can be obtained at other levels. 
The distribution changes after pairing the Guide with a new Learner, and gradually becomes less spiky as learning progresses.
We refer to the supplementary materials for Figure A.1, an illustration of this distribution in level PutNextLocal, and Figure A.2, which shows both distributions in level GoToObj. 

%
%
%



\subsection{Input/messages correlations}

We now look at the conditional distribution of the Guide's messages, given the observational input. To do so, we categorize the agent's field of perception according to four salient observable facts: whether the target object lies directly ahead of the agent, on her right, on her left, or whether the target object is not currently visible. 


In Figure A.2, the results of the pretrained Guide are shown for level GoToObj. They show that the Guide has a clear preference for a particular message when the object that has to be navigated towards lies ahead, on the left or on the right. Apparently, the guidance message distribution is conditioned on the observation of such situational circumstances. When the target object is invisible, the distribution is more spread out, indicating that the agent lets her policy depend on other factors in such cases. 

Similar plots can be obtained at other levels, as shown for GoToLocal in Figure A.3 in the supplementary materials. 
Distributions at higher levels tend to be somewhat more diffuse, indicating a higher uncertainty about the messages to emit. This is probably due to the presence of distractors. Moreover, there must be many other observable facts in the environment that impact the distributions, and that are not included here.

\begin{figure}[h]
\begin{subfigure}{0.5\textwidth}
\includegraphics[width=\textwidth]{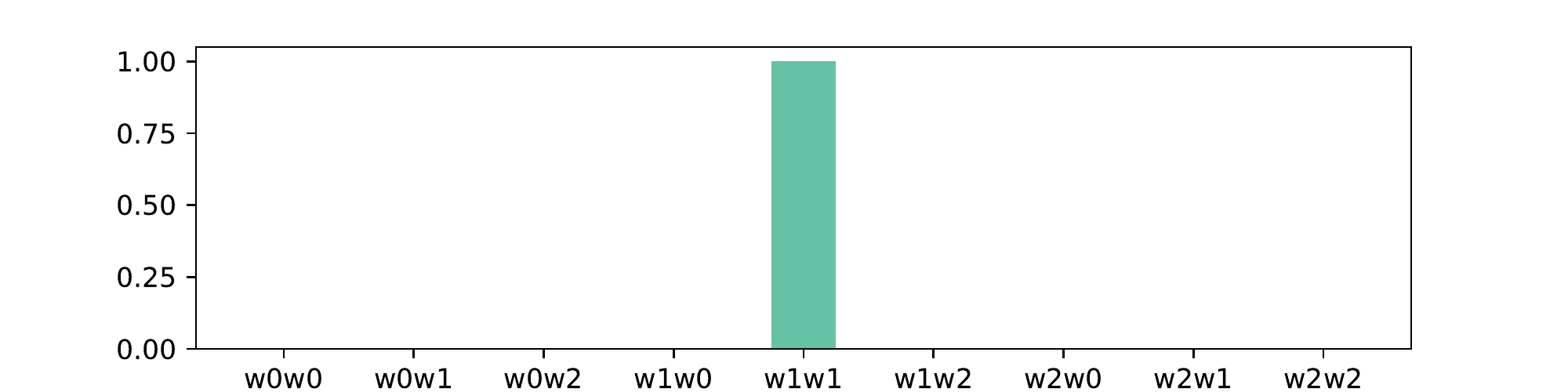}
\caption{Target object in front of agent.}
\end{subfigure}
\begin{subfigure}{0.5\textwidth}
\includegraphics[width=\textwidth]{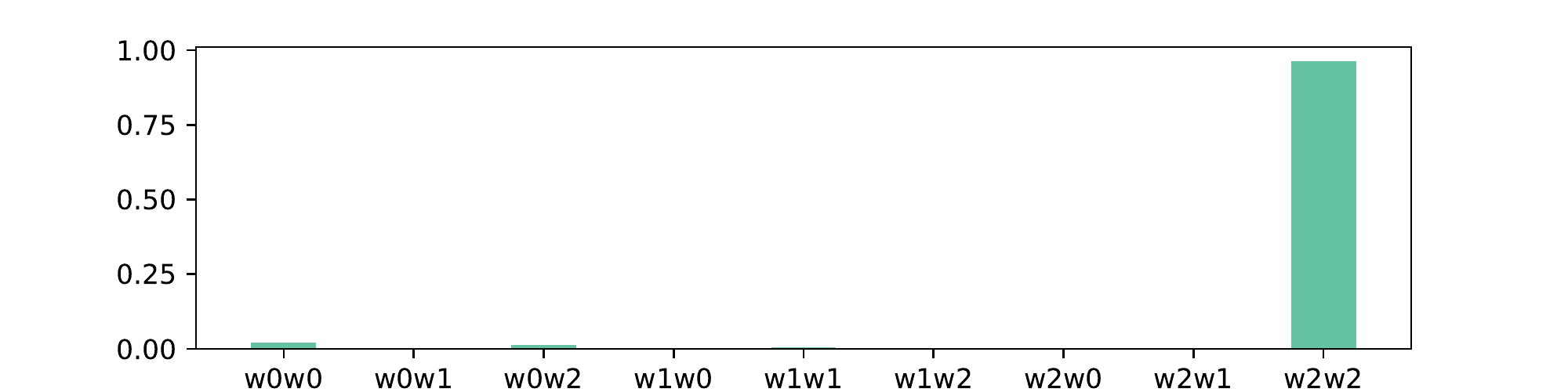}
\caption{Target object left of agent.}
\end{subfigure}
\begin{subfigure}{0.5\textwidth}
\includegraphics[width=\textwidth]{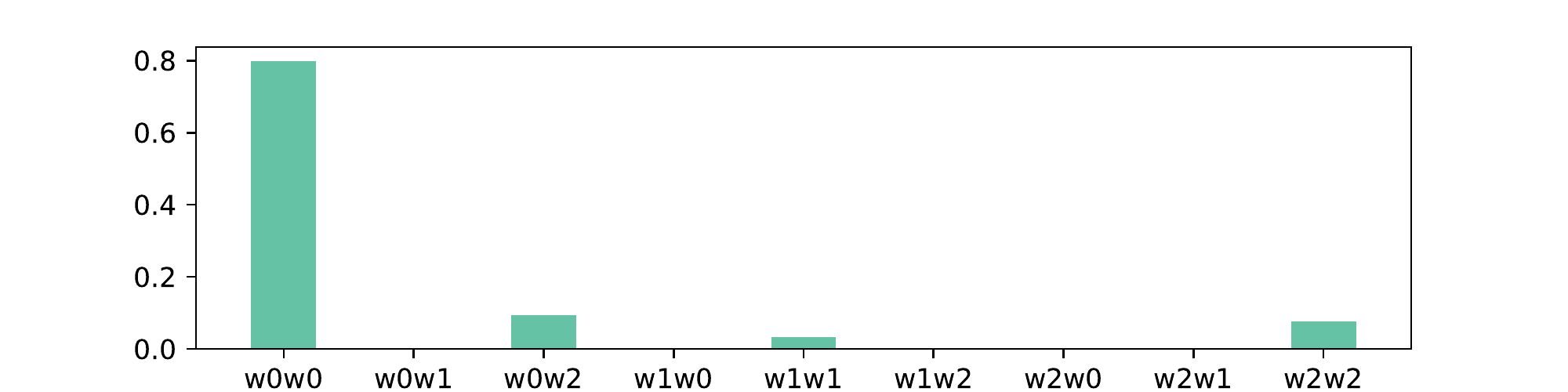}
\caption{Target object right of agent.}
\end{subfigure}
\begin{subfigure}{0.5\textwidth}
\includegraphics[width=\textwidth]{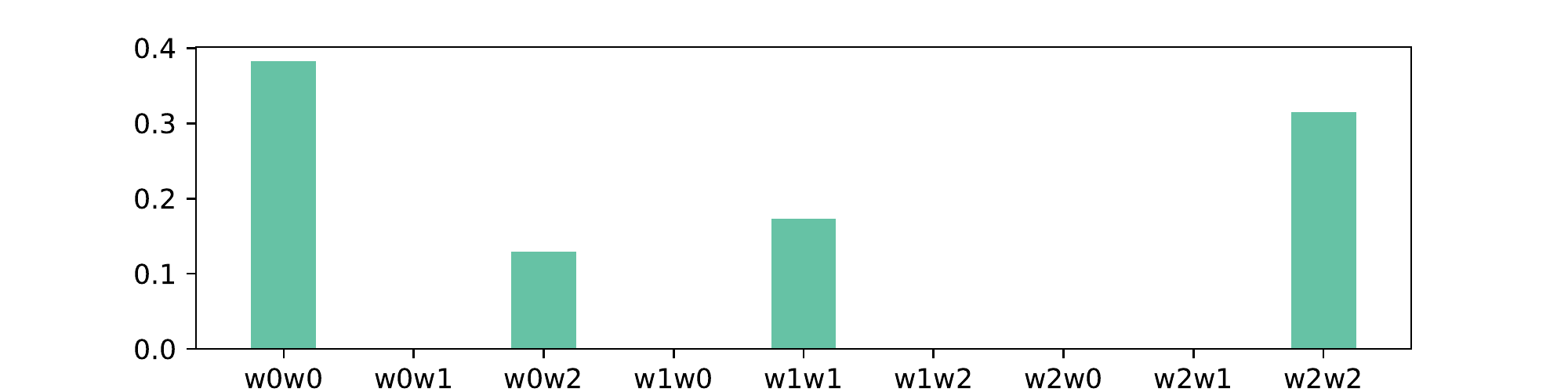}
\caption{Target object invisible.}
\end{subfigure}
\caption{Barplots visualizing the conditional distribution of messages produced by the best Guide in stage 2 at level GoToObj, given the indicated observable events, based on 500 sample episodes unseen during training.}
\label{fig:message-given-input-gotoobj}
\end{figure}


\subsection{Causal influence}

In the analysis so far, we looked at statistical correlations. This does not tell us anything yet about causality: do the Guide's messages actually \textit{cause} behavioral patterns in the Learner? To quantify the impact that the guidance messages have on the actions taken by a new Learner, we implement and compute a metric introduced for this purpose by \citet{lowe2019pitfalls}: Causal Influence of Communication (CIC).

CIC is intended to measure `positive listening': the extent to which emergent communication influences an agent's behavior. We consider one-step causal influence, c.q. the effect of the Guide's message on the next action of the Learner. It is computed as:

\begin{equation}
\text{CIC} = \dfrac{1}{\mid T \mid} \sum_{t \in T} \sum_{m \in M} \sum_{a \in A} p_t(a,m) \log \frac{p_t(a,m)}{p_t(a)p_t(m)},
\end{equation}

where $T$ is the set of `test games' (c.q. the frames of the validation episodes), $M$ the set of messages the Guide can send and $A$ the set of actions the Learner can take. Each $t \in T$ determines the probability distributions $p_t(a)$, $p_t(m)$ of Learner's actions and Guide's messages, as well as their joint distribution $p_t(a,m)$.

\begin{figure}[t]
\centering
\begin{subfigure}{0.23\textwidth}
\includegraphics[width=\textwidth]{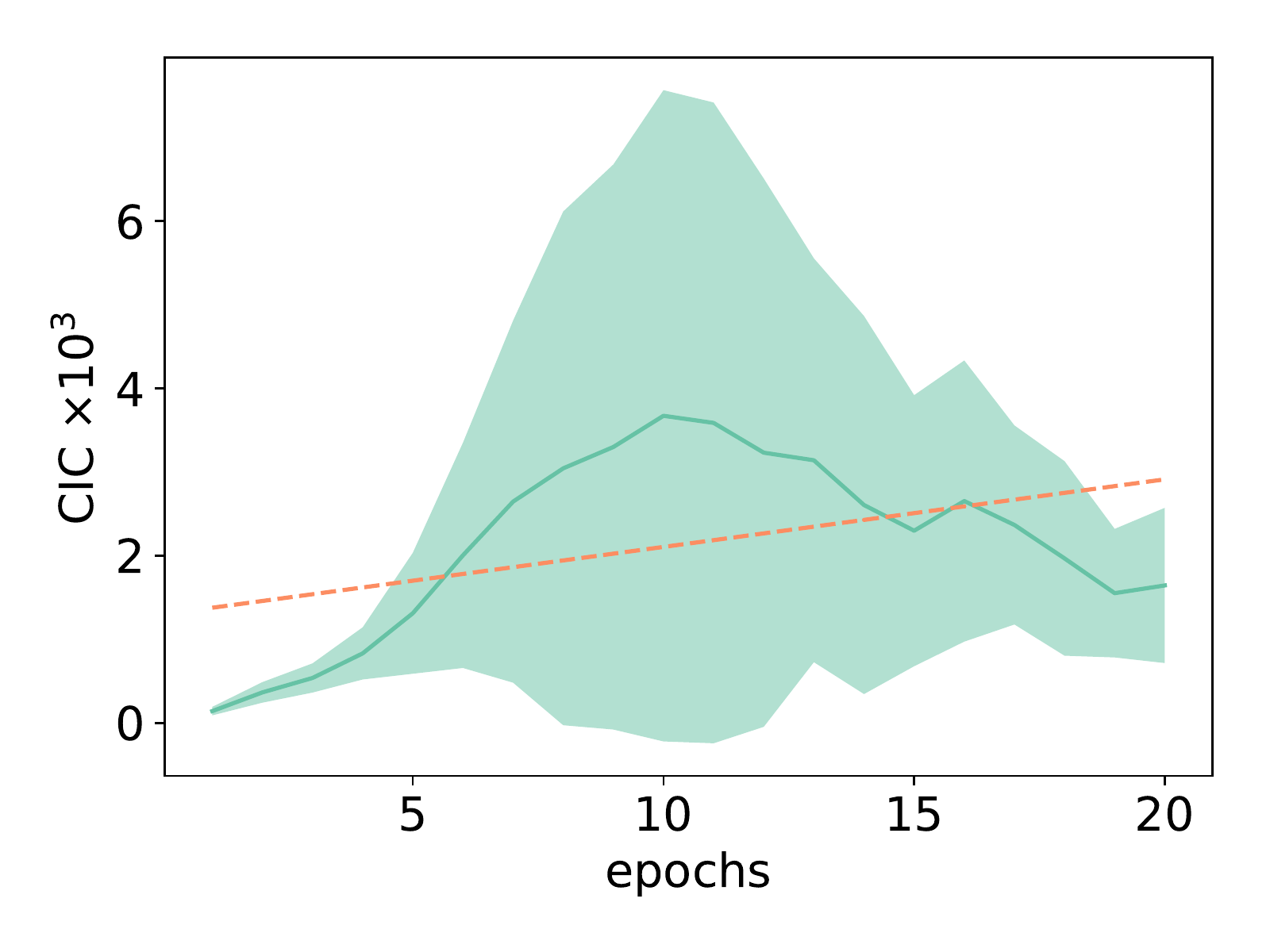}
\caption{GoToObj}
\end{subfigure}
\begin{subfigure}{0.23\textwidth}
\includegraphics[width=\textwidth]{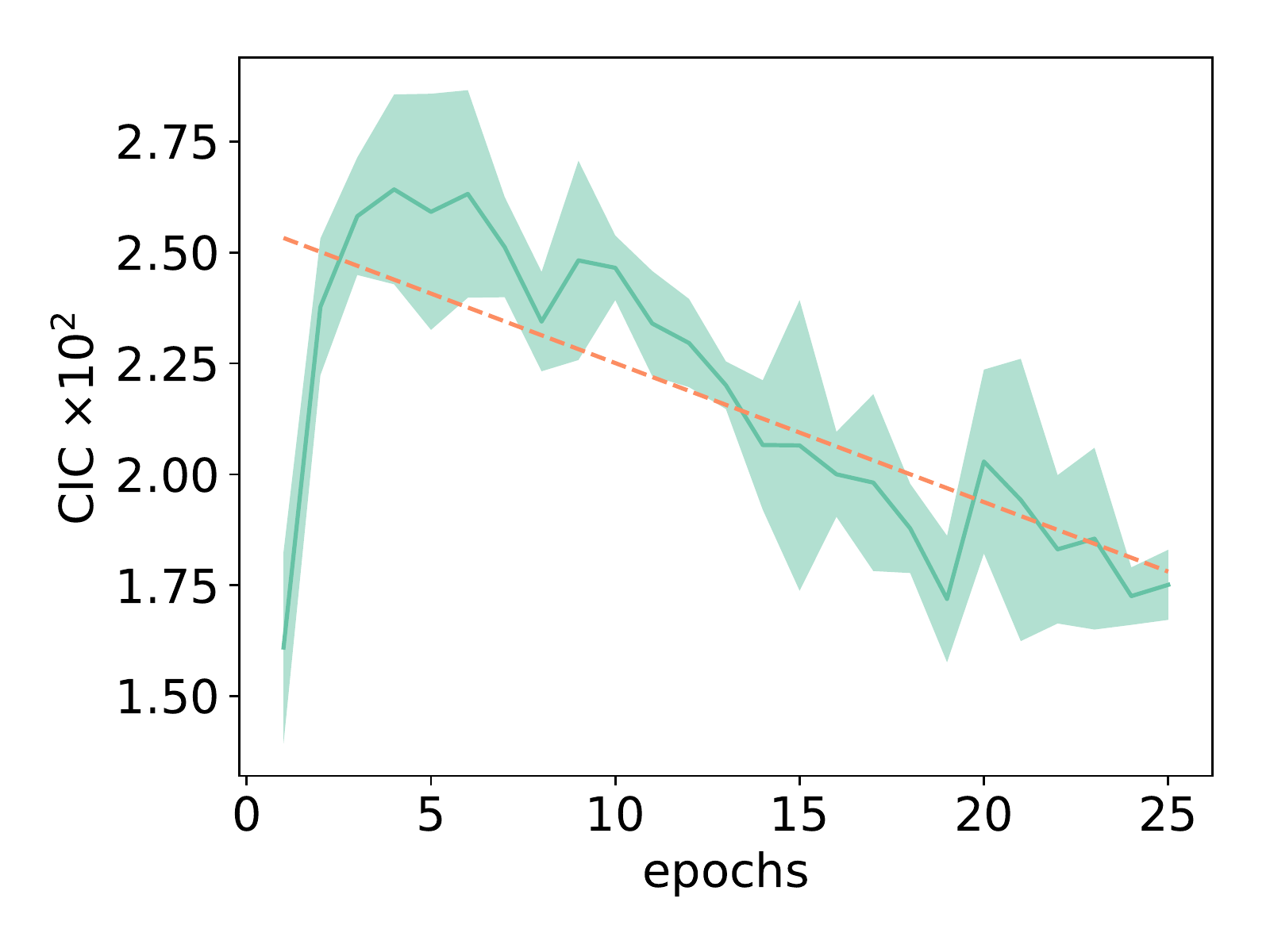}
\caption{GoToLocal}
\end{subfigure}
\begin{subfigure}{0.23\textwidth}
\includegraphics[width=\textwidth]{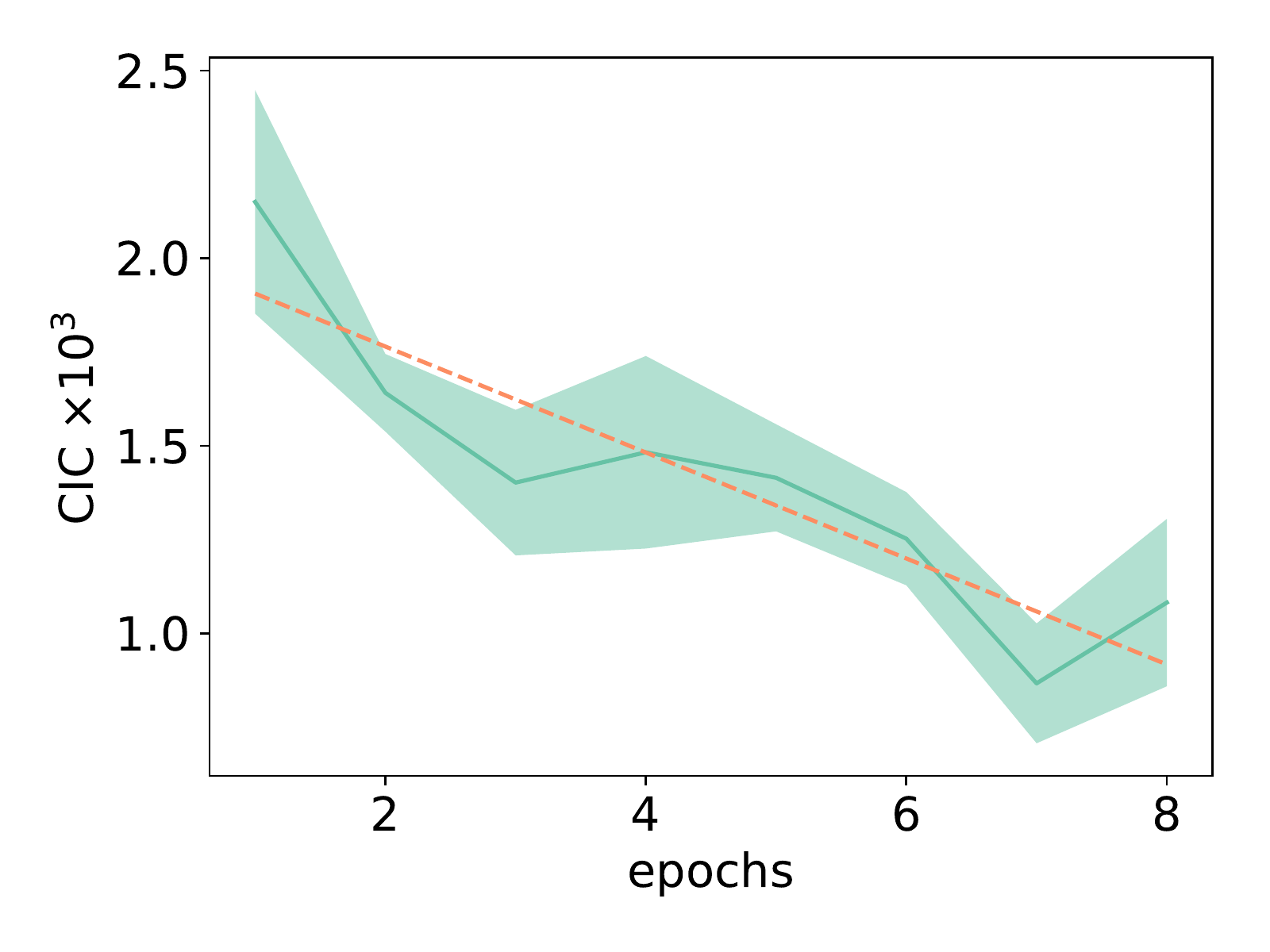}
\caption{GoToObjMaze}
\end{subfigure}
\begin{subfigure}{0.23\textwidth}
\includegraphics[width=\textwidth]{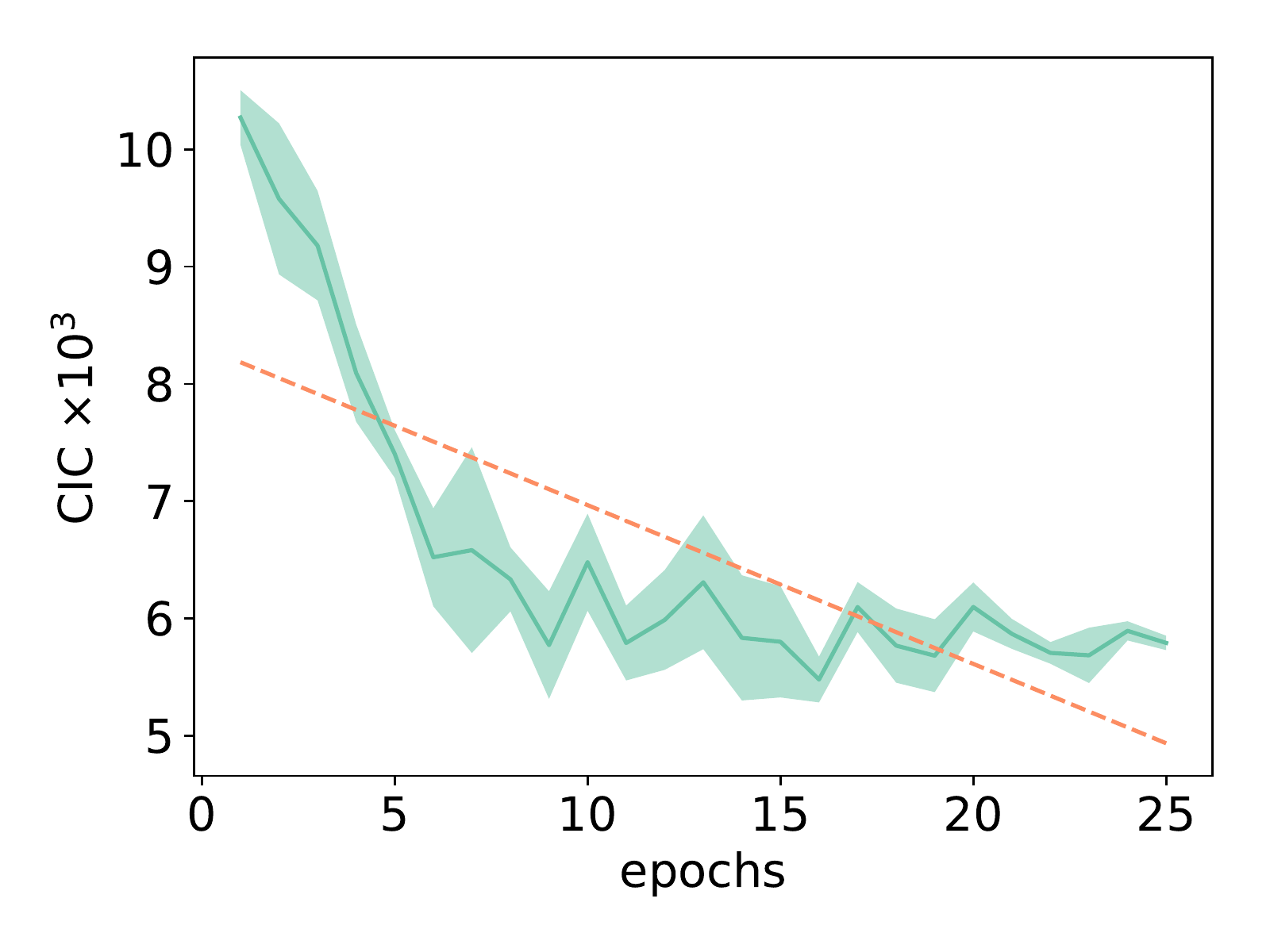}
\caption{PickupLoc}
\end{subfigure}
\begin{subfigure}{0.23\textwidth}
\includegraphics[width=\textwidth]{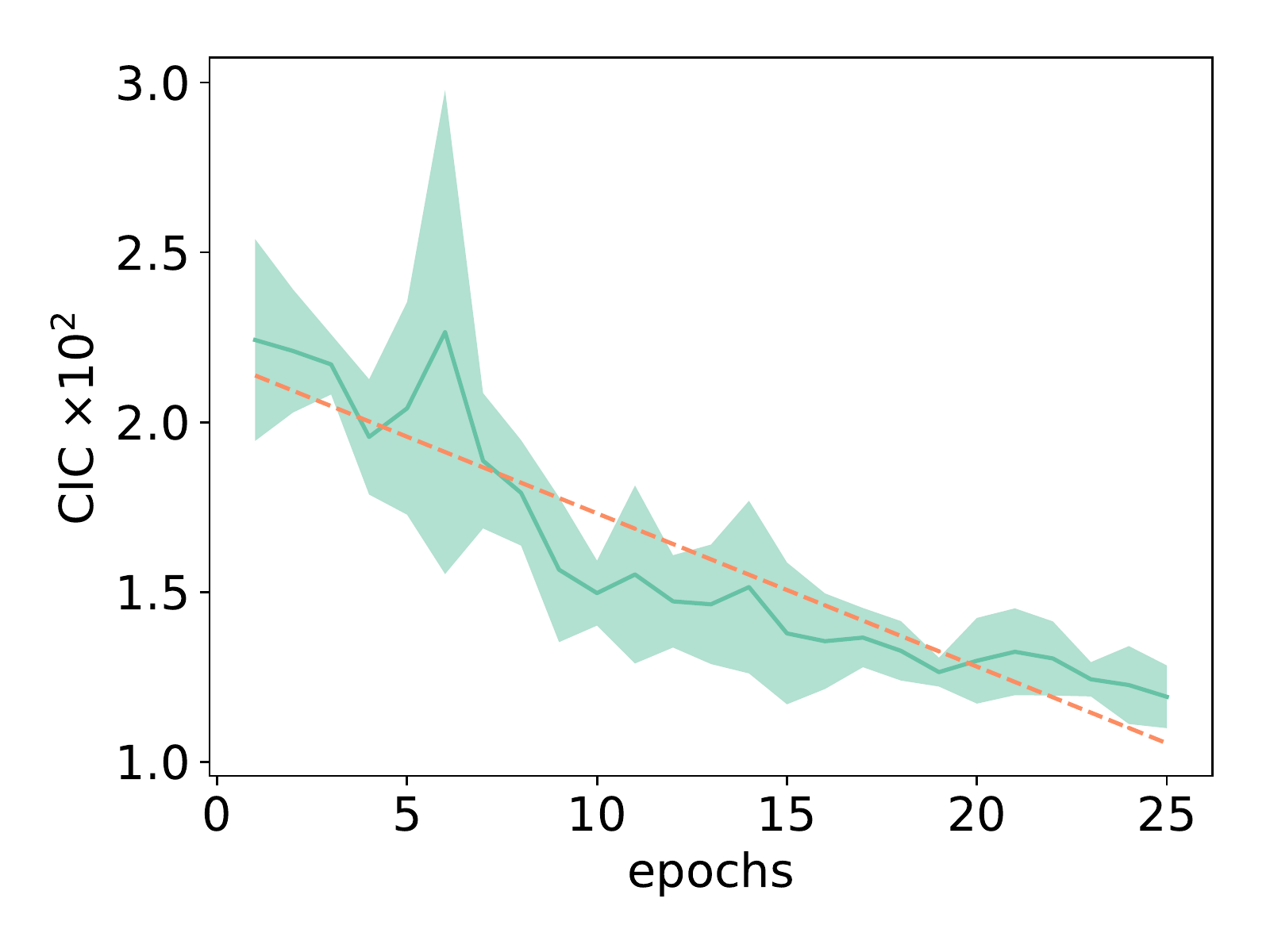}
\caption{PutNextLocal}
\end{subfigure}
\begin{subfigure}{0.23\textwidth}
\includegraphics[width=\textwidth]{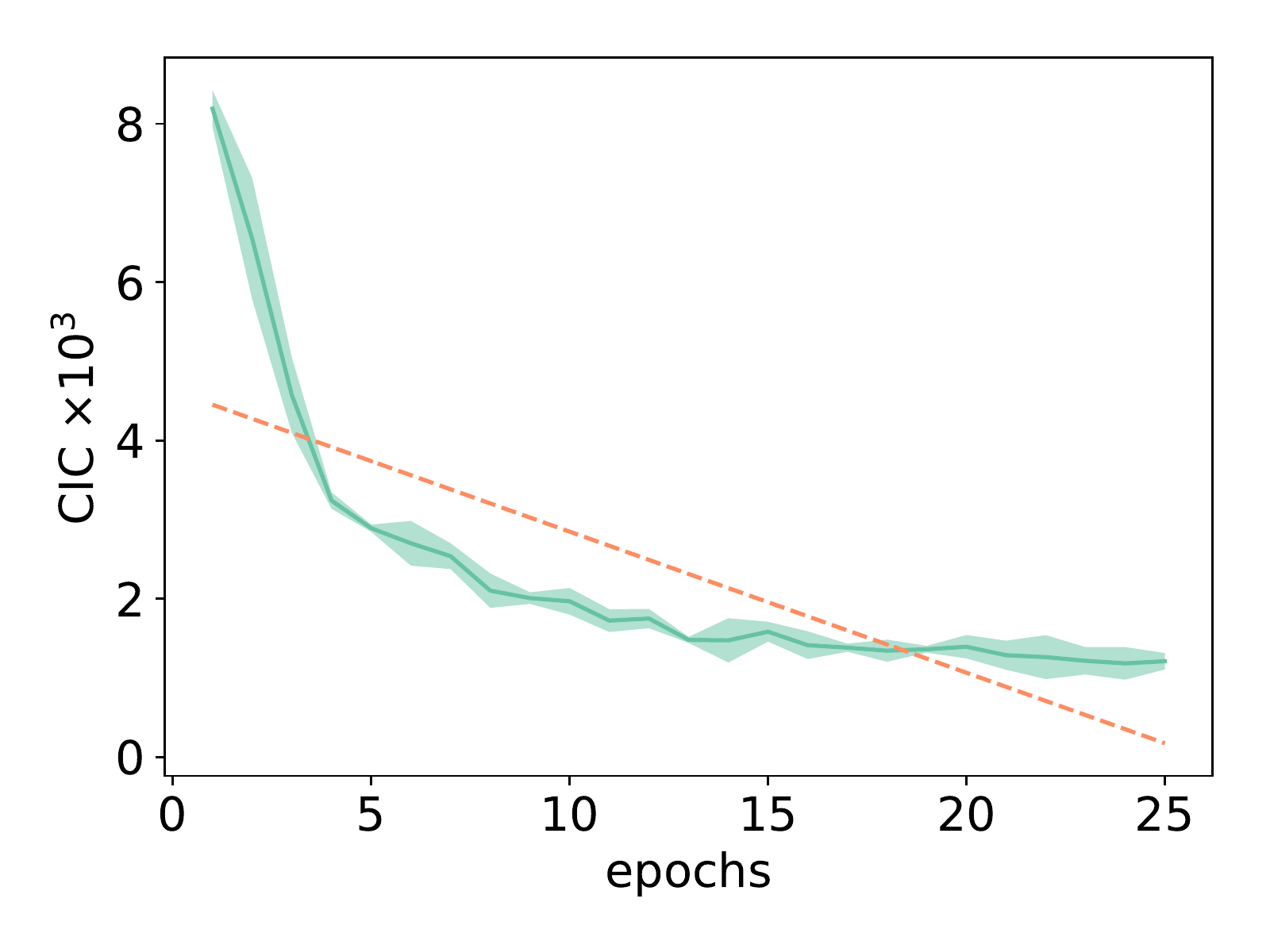}
\caption{GoTo}
\end{subfigure}
\caption{Development of CIC on 500 validation samples when training a Learner assisted by a pretrained Guide on the indicated BabyAI levels. Average over three runs; standard deviations are shown by shaded regions. Dashed orange lines show linear trends.}
\label{fig:cic-all-levels}
\end{figure}

Figure \ref{fig:cic-all-levels} visualizes the development of the CIC metric in the set-up combining a new Learner with a pretrained Guide at all considered levels. The plots clearly show that CIC tends to decrease over time, as illustrated by the linear trendlines. This indicates that the guidance messages have a high causal effect on the Learner's actions in the early stages, and gradually become less important as training progresses and the Learner grows more confident. The only exception in Figure \ref{fig:cic-all-levels} is level GoToObj, where we see a positive correlation between training time and CIC. We note, however,  that also at this level the trend is negative if more epochs after model convergence are considered.


\subsection{Learning the agents' language}

The correlations between actions and messages taught us how to interpret the Guide's language. 
This means that we do not only know how to make sense of witnessed utterances, but that we can also try to speak the language ourselves, encouraging the Learner to perform the actions that have proven to correlate with particular messages. 
To test this, we train a Learner with a pretrained Guide at levels GoToObj and GoToLocal, and identify the epoch where the CIC is highest. 
This should be the point in time where the Learner is most susceptible to the message channel. 
We compute the correlations between guidance messages and performed actions at this stage, as in Section \ref{sec:messages-actions}, to establish the expected correspondence between messages and actions. 
Next, we `hijack' the Guide, and start sending our own messages to the Learner. 

Communication success is assessed in two ways. 
First, we provide the Learner with a set of messages that should describe a pre-set trajectory (a `choreography') in the grid, as per the computed message-action correlations. In particular, we try to tell the agent to turn around its own axis (the `pirouette'), and to describe a larger circular movement (the `waltz'). 
We check if the performed action sequences adhere to these choreographies. 
Second, we report quantitative results by randomly sending the Learner 500 messages that strongly correlate with specific actions. 
We then calculate the `obedience': the percentage of messages that are followed by the expected action.

\begin{figure}[h]
\begin{subfigure}{0.23\textwidth}
\includegraphics[width=\textwidth]{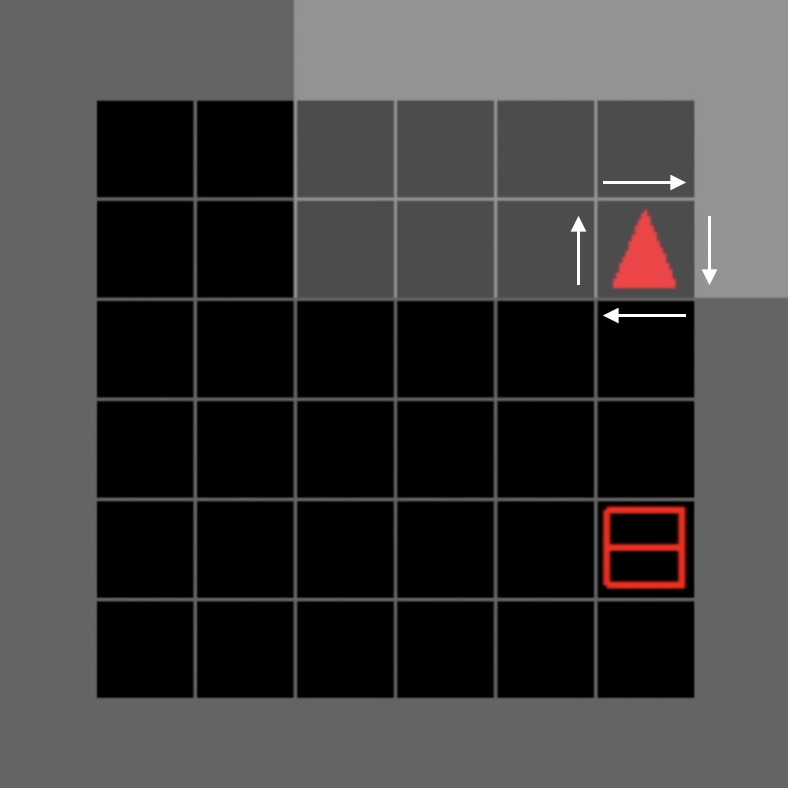}
\caption{Pirouette in GoToObj}
\label{fig:gotoobj-pirouette}
\end{subfigure}
\begin{subfigure}{0.23\textwidth}
\includegraphics[width=\textwidth]{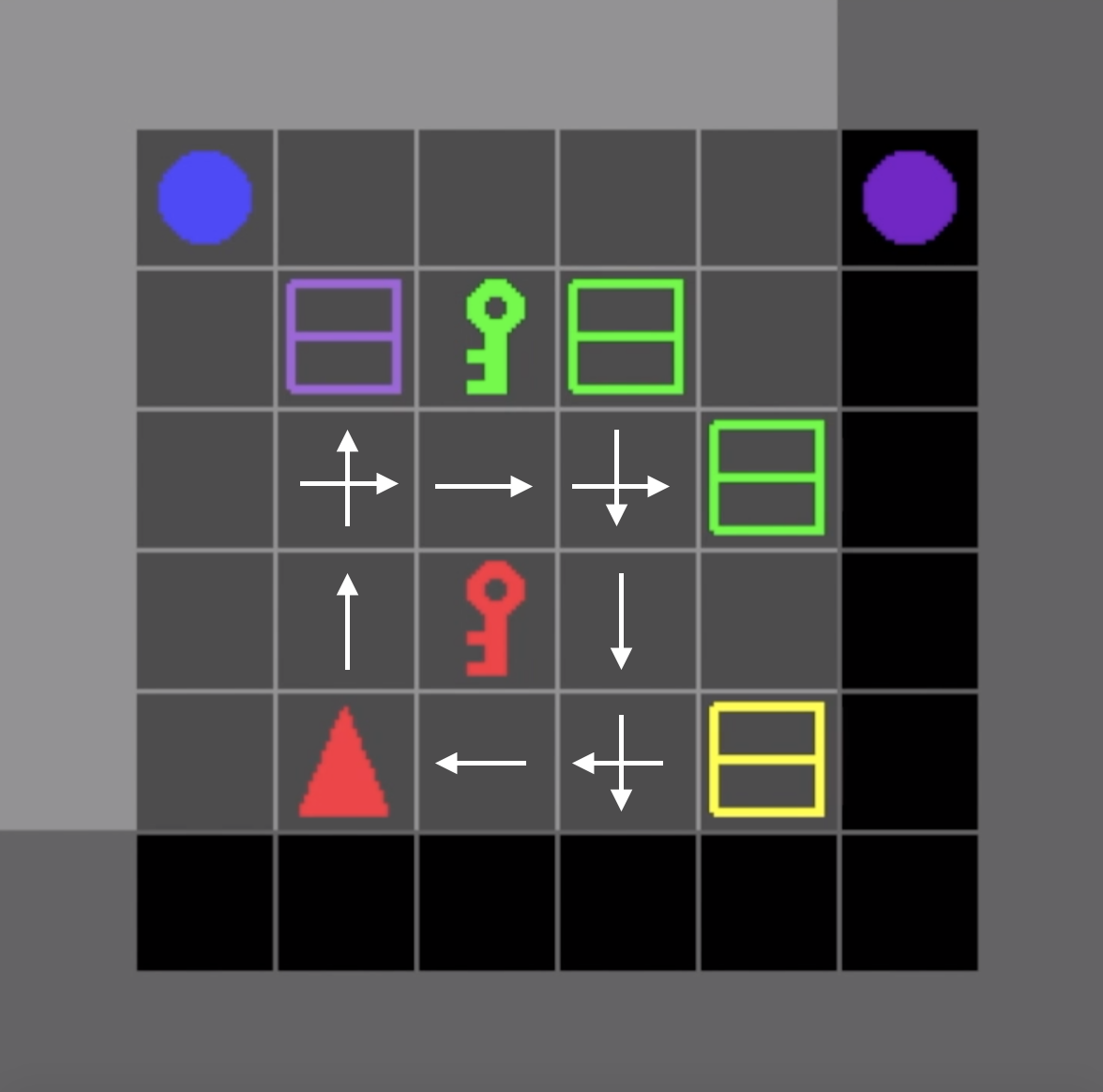}
\caption{Waltz in PutNextLocal}
\label{fig:gotolocal-waltz}
\end{subfigure}
\caption{Trajectories of Learners in indicated levels when told by means of messages to perform (a) a pirouette (c.q. $4 \times \texttt{w0 w0}$) and (b) a waltz (c.q. $4 \times \texttt{w1 w2}, \texttt{w1 w2}, \texttt{w1 w1}$ )}
\label{fig:choreographies}
\end{figure}

In GoToObj, the Learner successfully performs a pirouette and a waltz upon receiving the guidance messages that are supposed to describe these movements. See Figure \ref{fig:gotoobj-pirouette} for the pirouette. This is also the case in GoToLocal. See Figure \ref{fig:gotolocal-waltz} for the waltz. In GoToObj we reach an obedience of 76\%, and in GoToLocal one of 72\%. 
These percentages show that the Learner does not blindly follow the guidance messages, even when the CIC is highest. 
The observational input always remains important, especially when random messages are sent so that the correlation between input and guidance is broken. 
However, in most cases the agent does what we tell her to. Hence, to some extent we have mastered the agent language. 


%
%
%
%

\section{Conclusion}

We showed how neural agents can autonomously develop a discrete communication channel to help each other solve the BabyAI game. 
By applying curriculum learning, we demonstrated that the guidance messages encode information that is general enough to be useful in unseen levels. 
In the analysis, we saw that messages correlate with actions as well as with observations. 
Application of the CIC metric revealed that new agents gradually become more independent of the guidance messages they receive. 
Finally, we used the interpretability of the emerged language to successfully communicate with agents directly. 

In future research we would like to extend the developed method to frameworks other than BabyAI. 
We also wish to investigate methods to break the strong correlation between messages and actions, and incentivize guidance of a more general nature. 
Moreover, we are interested in experimenting with the number of tokens, the message length and the learning method, to see if similar results can be obtained if we use reinforcement instead of imitation learning. 


\bibliography{emnlp2019}
\bibliographystyle{acl_natbib}

\clearpage

\end{document}


\appendix
\renewcommand\thefigure{\thesection.\arabic{figure}}    
\renewcommand\thetable{\thesection.\arabic{table}}    

\section{Supplemental Materials}
\label{sec:supplemental}

\begin{table}[h]
\begin{small}
\begin{tabular}{@{}lcccccccc@{}}
\toprule
             & \multicolumn{8}{c}{required skills}                         \\ \midrule
             & ROOM & DISTR-BOX & DISTR & GOTO & PUT & PICKUP & LOC & MAZE \\
GoToObj      & x    &           &       &      &     &        &     &      \\
GoToLocal    & x    & x         & x     & x    &     &        &     &      \\
GoToObjMaze  & x    &           &       &      &     &        &     & x    \\
PickupLoc    & x    & x         & x     &      &     & x      & x   &      \\
PutNextLocal & x    & x         & x     &      & x   &        &     &      \\
GoTo         & x    & x         & x     & x    &     &        &     & x    \\ \bottomrule
\end{tabular}
\end{small}
\caption{Considered BabyAI levels and skills required to solve them, from \citet{chevalier2018babyai}. ROOM: navigating a $6\times 6$ room; DISTR-BOX: ignoring distracting grey boxes; DISTR: ignoring distracting objects of any kind; GOTO: understanding `go to' instructions; PUT: understanding `put' instructions; PICKUP: understanding `pick up' instructions; LOC: understanding location expressed relative to initial agent position; MAZE: navigating a $3 \times 3$ maze of randomly connected $6 \times 6$ rooms.}
\label{tab:babyai-levels}
\end{table}

\begin{table*}[h]
\begin{tabular}{@{}ll@{}}
\toprule
\# observation CNN output units & 256                                                               \\
\# memory LSTM units            & 2048                                                              \\
\# instruction GRU units        & 256                                                               \\
word embedding size             & 256                                                               \\
guidance message length         & 2                                                                 \\
guidance vocabulary size        & 3                                                                 \\
batch size                      & 128 (GoToObj, GoTo), 256 (GoToObjMaze), \\
& \hspace{0.5cm} 512 (otherwise)\\           
learning rate                   & $1\cdot 10^{-4}$                                                  \\
optimizer                       & Adam ($\beta_1 = 0.9, \beta_2 = 0.999, \epsilon = 1\cdot10^{-5}$) \\ 
\# training instances & 1K (GoToObj), 25K (GoToObjMaze), 50K (GoToLocal), \\
& \hspace{0.5cm}  100K (otherwise) \\
\bottomrule
\end{tabular}
\caption{Hyperparameter settings used in experiments.}
\label{tab:hyperparams}
\end{table*}

\begin{figure}[h]
\centering
\includegraphics[width=0.8\textwidth, trim={0 0 0 1cm}, clip]{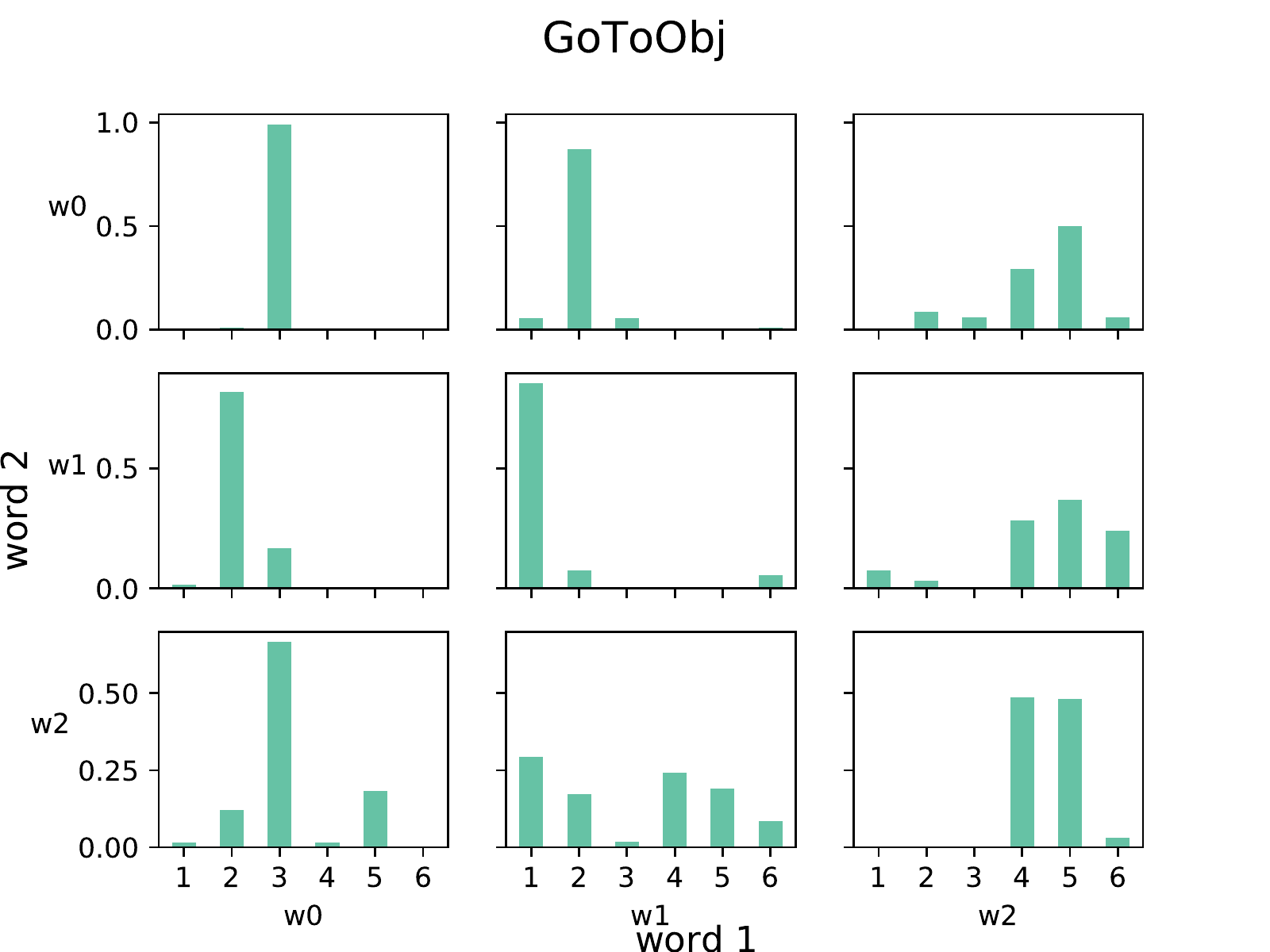}
\caption{Barplots visualizing the conditional distribution of actions taken in PutNextLocal by the best Learner after stage 3, given the Guide's messages, based on 500 sample episodes unseen during training. The bins in the individual barplots correspond to actions and the bar heights to probabilities. E.g. the top left barplot shows that the Guide performs action 3 with probability close to 1.0 after emitting message \texttt{w0 w0}.}\label{fig:action-message-dist-stage3-putnextlocal}
\end{figure}

\begin{figure}[h]
\centering
\begin{subfigure}{0.8\textwidth}
\includegraphics[width=\textwidth, trim={0 0 0 1cm}, clip]{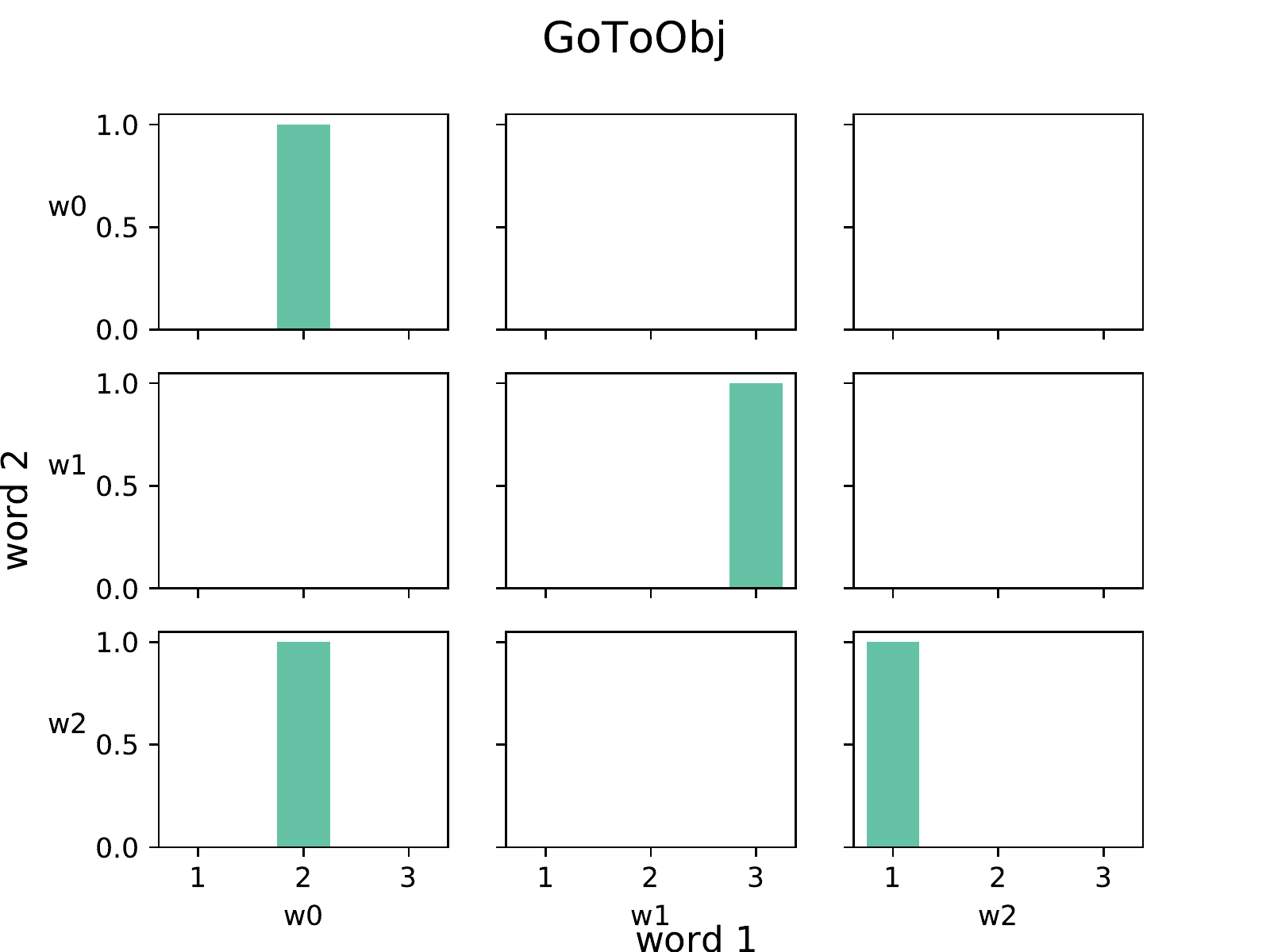}
\caption{}
\label{fig:action-message-dist-stage2-gotoobj}
\end{subfigure}
\begin{subfigure}{0.8\textwidth}
\includegraphics[width=\textwidth, trim={0 0 0 1cm}, clip]{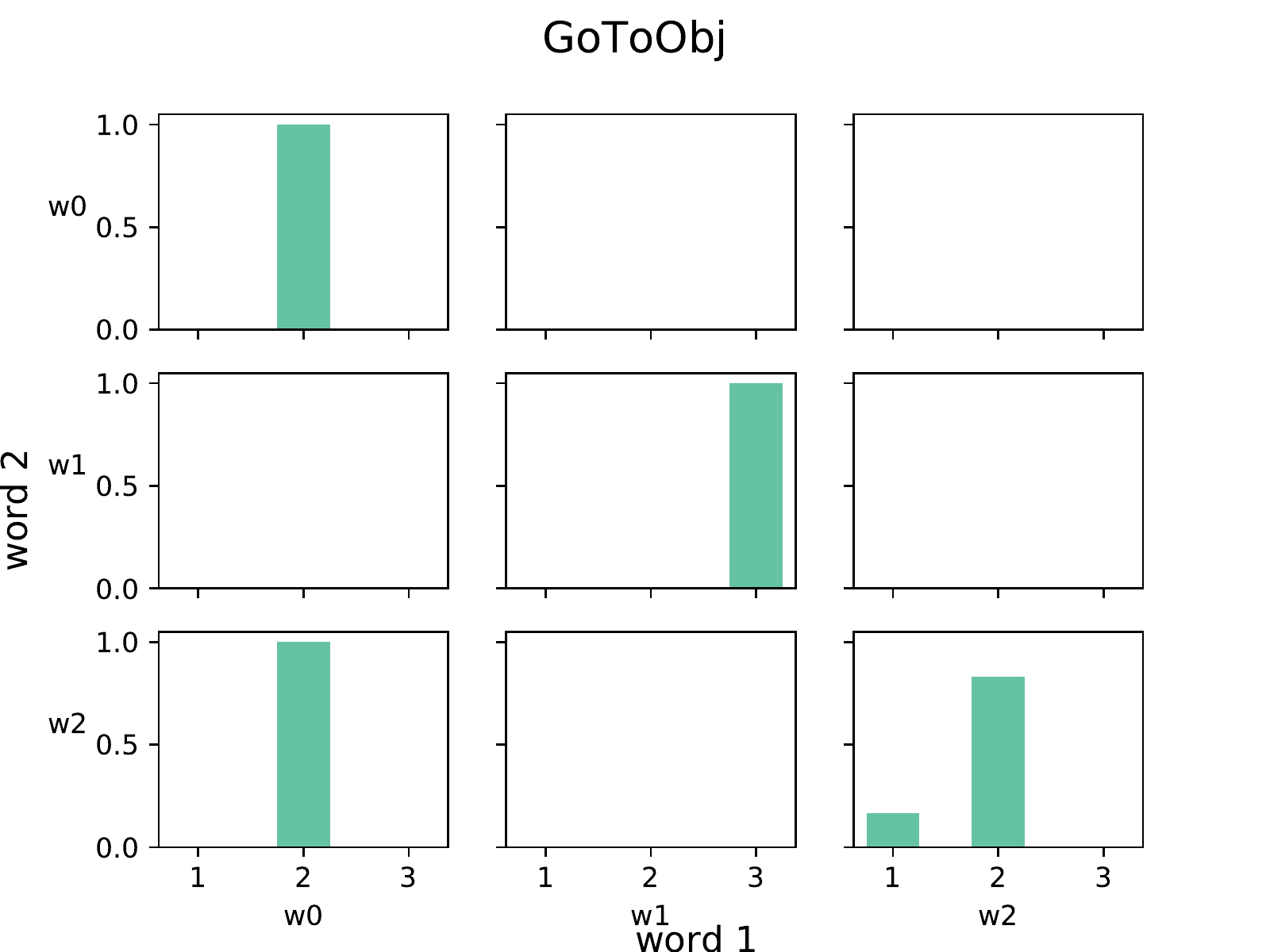}
\caption{}
\label{fig:action-message-dist-stage3-gotoobj}
\end{subfigure}
\caption{Barplots visualizing the conditional distribution of actions taken in GoToObj by (a) the best Guide after stage 2 and (b) the best Learner after stage 3, given the Guide's messages, based on 500 sample episodes unseen during training. The bins in the individual barplots correspond to actions and the bar heights to probabilities. E.g. the top left barplot in Figure (a) shows that the Guide performs action 2 with probability 1.0 after emitting message \texttt{w0 w0}.}
\label{fig:action-message-dist-gotoobj}
\end{figure}

\begin{figure}[h]
\centering
\begin{subfigure}{\textwidth}
\includegraphics[width=\textwidth]{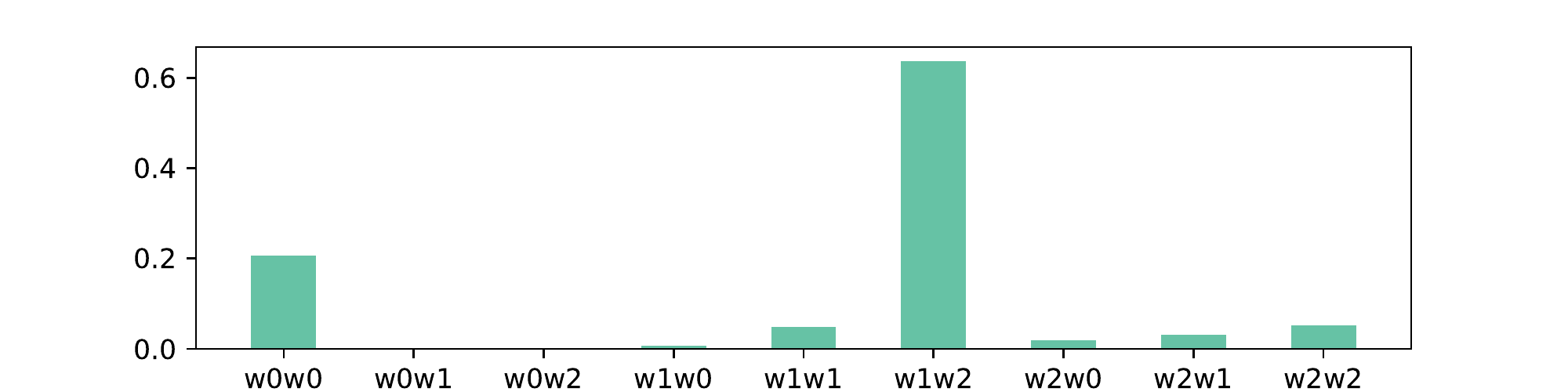}
\caption{Target object in front of agent.}
\end{subfigure}
\begin{subfigure}{\textwidth}
\includegraphics[width=\textwidth]{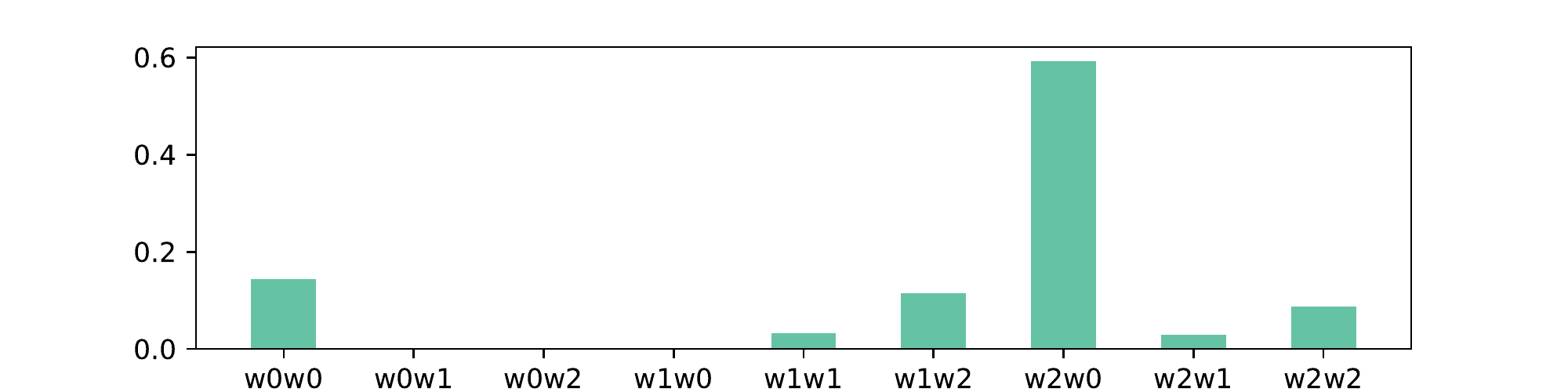}
\caption{Target object left of agent.}
\end{subfigure}
\begin{subfigure}{\textwidth}
\includegraphics[width=\textwidth]{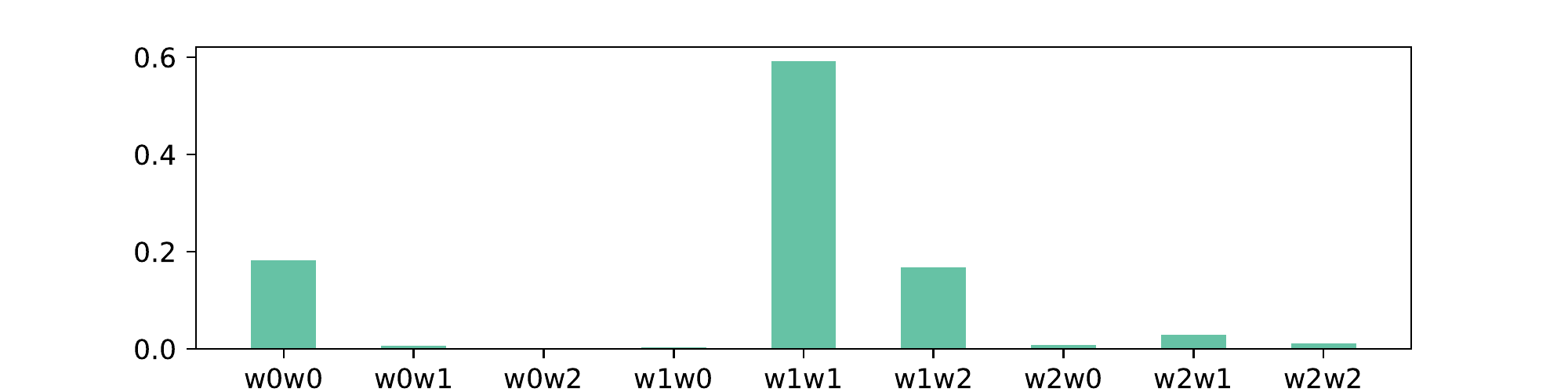}
\caption{Target object right of agent.}
\end{subfigure}
\begin{subfigure}{\textwidth}
\includegraphics[width=\textwidth]{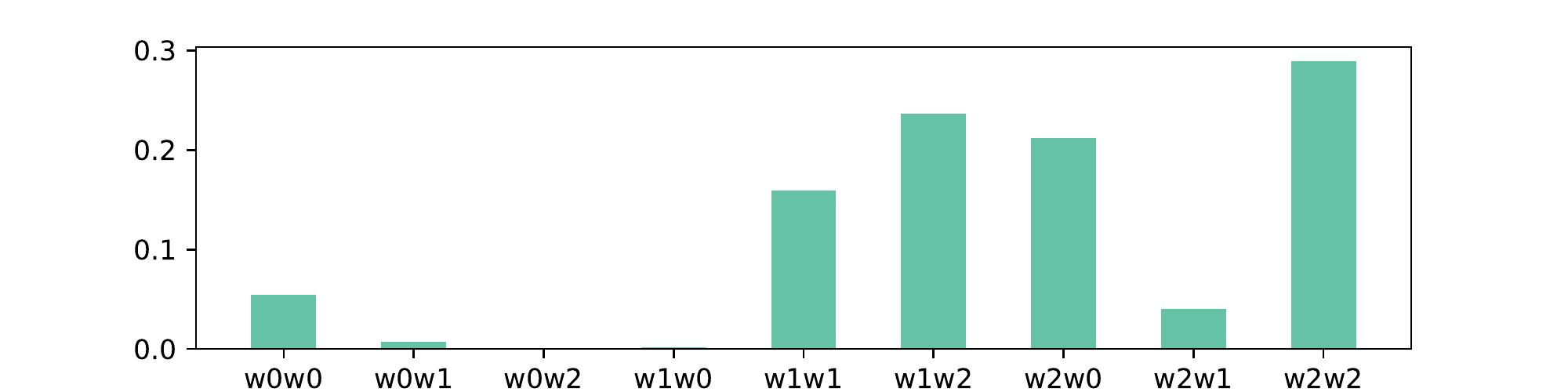}
\caption{Target object invisible.}
\end{subfigure}
\caption{Barplots visualizing the conditional distribution of messages produced by the best Guide in stage 2 at level GoToLocal, given the indicated observable events, based on 500 sample episodes unseen during training.}
\label{fig:message-given-input-gotolocal}
\end{figure}

\nobibliography{emnlp2019}
\bibliographystyle{acl_natbib}